\documentclass{article}

\usepackage[preprint]{neurips_2025}

\usepackage{times}

\usepackage{amsmath}
\usepackage{listings}
\usepackage{mdframed}
\usepackage{enumitem}
\usepackage{float}
\usepackage{titlesec}

\usepackage{xcolor}

\usepackage{tabularx}
\usepackage{amssymb}
\usepackage{xr} 
\usepackage{graphicx}
\graphicspath{ {./images/} }

\usepackage[utf8]{inputenc} 
\usepackage[T1]{fontenc}
\usepackage{hyperref} 

\usepackage[capitalize,noabbrev]{cleveref}

\usepackage{url}
\usepackage{booktabs} 
\usepackage{amsfonts} 
\usepackage{nicefrac}   
\usepackage{microtype} 
\usepackage{xcolor} 

\makeatletter
\newcommand*{\addFileDependency}[1]{
\typeout{(#1)}
\@addtofilelist{#1}
\IfFileExists{#1}{}{\typeout{No file #1.}}
}
\makeatother

\newcommand*{\myexternaldocument}[1]{%
\externaldocument{#1}%
\addFileDependency{#1.tex}%
\addFileDependency{#1.aux}%
}

\usepackage[para,multiple]{footmisc}

\myexternaldocument{supplemental}

\title{CogniLoad: A Synthetic Natural Language Reasoning Benchmark With Tunable Length, Intrinsic Difficulty, and Distractor Density}

\author{Daniel Kaiser\thanks{Integreat - Norwegian Centre for knowledge-driven machine learning}~~\thanks{UiT - The Arctic University of Norway}~~ Arnoldo Frigessi\footnotemark[1]~~\thanks{University of Oslo}~~ Ali Ramezani-Kebrya\footnotemark[1]~~\footnotemark[3]~~ Benjamin Ricaud\footnotemark[1]~~\footnotemark[2]\\ \\
$^{\dagger}$\href{mailto:daniel.kaiser@uit.no}{\texttt{daniel.kaiser@uit.no}}
}

\begin{document}
\maketitle

\begin{abstract}
Current benchmarks for long-context reasoning in Large Language Models (LLMs) often blur critical factors like intrinsic task complexity, distractor interference, and task length. To enable more precise failure analysis, we introduce \textbf{CogniLoad}, a novel synthetic benchmark grounded in Cognitive Load Theory (CLT). CogniLoad generates natural-language logic puzzles with independently tunable parameters that reflect CLT's core dimensions: intrinsic difficulty ($d$) controls intrinsic load; distractor-to-signal ratio ($\rho$) regulates extraneous load; and task length ($N$) serves as an operational proxy for conditions demanding germane load. Evaluating 22 SotA reasoning LLMs, CogniLoad reveals distinct performance sensitivities, identifying task length as a dominant constraint and uncovering varied tolerances to intrinsic complexity and U-shaped responses to distractor ratios. By offering systematic, factorial control over these cognitive load dimensions, CogniLoad provides a reproducible, scalable, and diagnostically rich tool for dissecting LLM reasoning limitations and guiding future model development.
\end{abstract}

%%%%%%%%%%%%%%%%%%%%%%%%%%%%%

\section{Introduction}

Cognitive Load Theory (CLT) \citep{sweller1988cognitive} characterizes three types of cognitive load on human working memory when solving problems \citep{sweller1988cognitive,paas2003cognitive,lieder2020resource}: intrinsic (ICL), extraneous (ECL), and germane (GCL). ICL stems from the inherent complexity and element interactivity of the task \citep{halford1998processing}. ECL is induced by suboptimal task presentation requiring the processing of elements that are not task-relevant \citep{chandler1991cognitive}. GCL concerns effective remaining resources allocated to engaging with the intrinsic task demands for mental schema construction  \citep{ericsson1995long,sweller2010element}.

Large language models (LLMs) demand analogous computational resources when solving reasoning tasks. The essential element interactivity of a reasoning chain mirrors ICL; distractor elements reflect ECL; and sustained engagement with intrinsically relevant information over a long reasoning process acts as an operational proxy for germane-like processing - the constructive effort to maintain a coherent problem representation.

To the best of our knowledge, no study has based the evaluation of problem-solving capacities of LLMs in CLT by distinguishing these three load types, and existing benchmarks often confound them: LongBench \citep{bai2024longbench} and L-Eval \citep{an2024eval} vary context length but not necessarily the intrinsic reasoning depth; LogicBench \citep{parmar2024logicbench} probes ICL with minimal demands on ECL or context-induced load; BABILong \citep{kuratov2024babilong} mixes multi-step reasoning with fixed distractor ratios, obscuring precise failure attribution.

We introduce \textbf{CogniLoad}, a controllable synthetic benchmark for long-context reasoning, inspired by CLT, that operationalizes these load types through tunable parameters in randomized natural-language logic puzzles:
    \textbf{(i) Intrinsic Load} via Intrinsic Difficulty $d$ controls the number of interacting entities, attributes, and logical clauses, directly manipulating ICL by varying essential element interactivity and reasoning depth.
    % \item 
    \textbf{(ii) Extraneous Load} via Distractor Density $\rho$l dictates distractor density; lower $\rho$ increases irrelevant elements, manipulating ECL.
    % \item 
    \textbf{ (iii) Germane Load Proxy} via Task Length $N$ serves as an operational proxy for demanding germane-like cognitive work.
% \end{enumerate}

\begin{figure}[t]
    \centering
    \includegraphics[width=\textwidth]{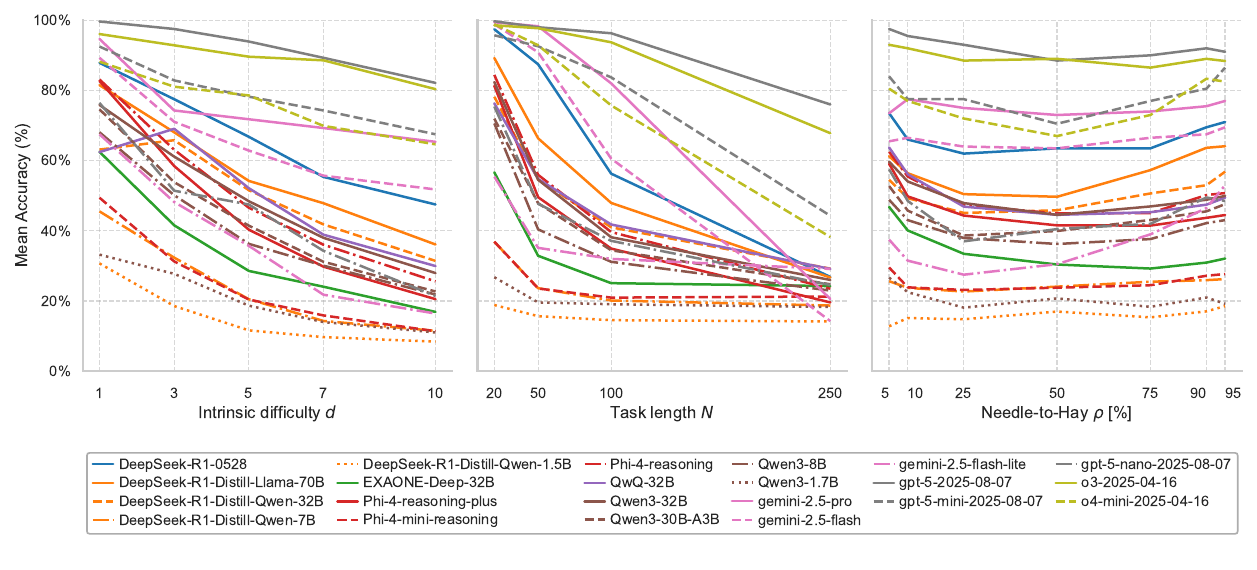}
    \caption{The average accuracy of models  across the evaluated parameter space for $d \in \{1,3,5,7,10\}$ (left panel), $N \in \{20,50,100,250\}$ (center panel),  and $\rho \in \{5,...,95\}$ (right panel). Each plot selects one dimension for the X-axis and averages the accuracy of all evaluated puzzles for the other two dimensions relative to it.}
    \label{fig:results}
\end{figure}

Our key contributions are summarized as follows:
\begin{enumerate}
\item  We ground the evaluation of LLMs in CLT, precisely defining benchmark parameters that control ICL, ECL, and an operational proxy for the conditions conducive to GCL.
\item We introduce \emph{CogniLoad}, the first benchmark designed to independently control these three dimensions of cognitive load, while scaling to arbitrarily long contexts.
\item  We provide an algorithm for the automatic randomized generation and evaluation of puzzle instances, enabling large-scale and reproducible comparison of LLM capabilities.
\item We report empirical results on 22 state-of-the-art (SotA) reasoning LLMs (see \cref{fig:results}), revealing distinct failure regimes across the \((d,N,\rho)\) dimensions and highlighting specific targets for improving LLM design. 
\end{enumerate}

Together, these contributions translate CLT into a precise diagnostic framework for understanding and advancing long-context reasoning in LLMs.

\subsection{Related work}

\textbf{Long-context Benchmarks (Working Memory Capacity).}
A line of work starting with Long-Range Arena (LRA) \citep{tay2020long} and followed by several recent benchmarks probe LLM performance on long sequences, often framed as testing ``memory load'' or context utilization. Earlier studies such as SCROLLS \citep{shaham2022scrolls}, BookSum \citep{kryscinski2021booksum}, and QMSum \citep{zhong2021qmsum} scale document length without manipulating intrinsic difficulty. LongBench \citep{bai2024longbench,bai2024longbenchv2} and L-Eval \citep{an2024eval} aggregate multi-task corpora up to 200k tokens, while BABILong \citep{kuratov2024babilong}, LongReason \citep{ling2025longreason}, RULER \citep{hsieh2024ruler}, ZeroSCROLLS \citep{shaham2023zeroscrolls}, and Michelangelo \citep{vodrahalli2024michelangelo} increase context while the inherent difficulty of individual sub-tasks (ICL) may vary unsystematically and distractor density (ECL) is often not a controlled variable. Consequently, performance degradation could be due to sheer length overwhelming processing capacity, or an inability to sustain germane-like cognitive work over extended relevant information, but the precise cause of failure is not clear.

\textbf{Logical-reasoning Benchmarks (Intrinsic Load).}
A complementary line of benchmarks focuses on ICL by presenting tasks with high inherent complexity but often within minimal context lengths or distractors. Notable classical suites include ReClor \citep{yu2020reclor}, LogiQA \citep{liu2020logiqa}, and BIG-Bench-Hard (BBH) \citep{suzgun2022challenging}. AutoLogic \citep{zhu2025autologi} is a benchmark that explicitly focuses on scaling ICL through controllable complexity. LogicBench \citep{parmar2024logicbench}, CLUTRR \citep{sinha2019clutrr}, and ZebraLogic \citep{lin2025zebralogic} also exemplify this by formulating symbolic logic puzzles that demand processing many interacting  elements (e.g., multi-step deductions, handling negation, and constraint satisfaction). Similarly, mathematical reasoning datasets, e.g., GSM8K \citep{cobbe2021training} and abstract rule induction tasks, e.g., ARC-AGI \citep{chollet2024arc} primarily escalate ICL by increasing the complexity of essential rules and their interdependencies. 

\textbf{Needle in a Haystack Benchmarks (Extraneous Load).}
Needle in a haystack (NIAH) designs \citep{gkamradt23} specifically target ECL by embedding relevant facts (``needles'') within large volumes of distractor text (``hay''). Variants such as Sequential NIAH \citep{yu2025sequential} and Nolima \citep{modarressi2025nolima} investigate the impact of such distractors, which constitute non-essential elements requiring processing for filtering, thereby imposing ECL. While these benchmarks effectively isolate the impact of distractors on information retrieval, the ``needle'' tasks themselves typically involve low ICL (e.g., simple fact lookup). 

\textbf{Need for Multi-dimensional Evaluation.}
CLT highlights the interplay of ICL, ECL, and germane processing under finite working memory \citep{paas2003cognitive}. Existing LLM reasoning benchmarks, however, typically manipulate only one dimension without systematic, independent control over the others. Even benchmarks like MIR-Bench \citep{yan2025mir} which combine high ICL with extensive input, do not offer the factorial control needed to disentangle these loads, hindering precise diagnostics. Similar to our work, GSM-$\infty$ \citep{zhou2025gsminfinitellmsbehaveinfinitely} allows manipulating noise and difficulty. However, these parameters are not adjusted independently of task length.

\textbf{Contribution of CogniLoad.} CogniLoad addresses this critical gap by providing a framework for independently controlling parameters that influence: (i) ICL via intrinsic puzzle difficulty ($d$), (ii) ECL via distractor density ($\rho$), and (iii) the demands for sustained, germane-like processing via task length ($N$), all within a single synthetically generated natural language puzzle. This factorial design enables a precise diagnosis of LLM failure modes, specifically the inability to handle increased intrinsic complexity, susceptibility to extraneous distractors, or incapacity to maintain coherent reasoning over an extended number of sequences. By explicitly grounding these dimensions in CLT, CogniLoad offers the first benchmark to diagnostically map LLM capability surfaces across these distinct cognitive demands, thereby complementing and extending the insights from evaluations that focus on a single factor.

\section{Benchmark Design: CogniLoad Logic Puzzles}
\label{sec:benchmark_design}

\subsection{Puzzle Definition and Construction}
\label{subsec:puzzle_definition}

CogniLoad is a family of natural-language logic-grid puzzles explicitly crafted to probe sequential reasoning capabilities of LLMs. The design goals are threefold: each puzzle (i) necessitates sequential multi-step deduction where order fundamentally matters; (ii) embeds a controllable number of relevant ``needle'' facts within the context of a controllable number of ``hay'' distractor statements; and (iii) provides parameters that control distinct dimensions of cognitive load. This section formalizes the task and describes the puzzle generation process, the control parameters, and key design choices.

Each puzzle in CogniLoad (see  ~\cref{fig:example-puzzle}) consists of a set of people with independent and mutable attributes. A series of statements, applied in strictly sequential order, updates these attributes according to  conditions specified in each statement. The puzzle generation is parameterized by three key parameters: intrinsic difficulty $d$, total number of statements $N$, and needle-to-hay ratio $\rho$.

\begin{figure}[t]
\begin{mdframed}[backgroundcolor=gray!10, linewidth=1pt]
\small
\textbf{(i) Puzzle Instruction:} Solve this logic puzzle. You MUST finalize your response with a single sentence about the asked property (e.g., "Peter is in the livingroom.", "Peter is wearing blue socks",.. ). Solve the puzzle by reasoning through the statements in a strictly sequential order.

\vspace{0.2cm}
\begin{minipage}[t]{0.48\textwidth}
\textbf{(ii) Initial State:}
\begin{itemize}[leftmargin=*,noitemsep,topsep=0pt]
\item Brent is wearing green socks and is wearing purple gloves and last listened to classical music.
\item Anthony is wearing purple socks and is wearing yellow gloves and last listened to disco music.
\item $\ldots$
\end{itemize}
\end{minipage}
\hfill
\begin{minipage}[t]{0.48\textwidth}
\textbf{(iii) Update Statements:}
\begin{enumerate}[leftmargin=*,noitemsep,topsep=0pt]
\item The people wearing green socks listen to electronic music.
\item The people who last listened to classical music and wearing purple gloves put on yellow gloves.
\item $\ldots$
\end{enumerate}
\end{minipage}

\vspace{0.2cm}
\hspace{0.25\textwidth}\textbf{(iv) Query:} What color of socks is Brent wearing?
\end{mdframed}

\caption{Example CogniLoad puzzle with intrinsic difficulty $d=3$, statements $N=20$, and needle-to-hay ratio $\rho=50\%$. Only a subset of the initial state and update statements is shown.}
\label{fig:example-puzzle}
\end{figure}

\subsubsection{Basic Puzzle Construction}

A puzzle is formally characterized by the following components:

\begin{itemize}
    \item \textbf{People}: A set $P = \{p_1, p_2, \ldots, p_n\}$ of persons in the puzzle,  and $n = \max(d, 2)$.
    \item \textbf{Person of Interest (PoI)}: A randomly selected person $p^* \in P$ about whom the final question is asked.
    \item \textbf{Attribute Categories}: A set $A = \{c_1, c_2, \ldots, c_d\}$ of attributes randomly selected from a predefined taxonomy of 12 categories. Each category takes values in a Value Domain with a given finite cardinality, smaller or equal to 10.
    \item \textbf{Value Domains}: For each category $c \in A$, a value domain $V_c = \{v_{c,1}, v_{c,2}, \ldots, v_{c,\ell_c}\}$ where $\ell_c = d+1$ for $d > 1$ or $\ell_c = 3$ when $d = 1$. See Table \ref{tab:ontology_snapshot} for the ontology.
    \item \textbf{State Function}: $S_t(p, c)$ represents the value of attribute $c$ for person $p$ at step $t$. Each person has values for the $d$ attribute of the selected attribute categories $A$, thus the state value represents a vector of dimension $d$.
\end{itemize}

\begin{table}[h!]
\centering
\caption{Overview of the attribute ontology. The full ontology contains 12 categories of varying domain sizes and is detailed completely in~\cref{supp-app:ontology}.} 
\label{tab:ontology_snapshot}
\begin{tabular}{lcl}
\toprule
Category Name (Code) & Domain Size & Examples of Values \\ \midrule
\texttt{location} & 50+ & kitchen, balcony, zoo, museum, park... \\
\texttt{clothes\_socks} & 10 & blue, red, yellow, green, purple... \\
\texttt{clothes\_gloves} & 10 & (same as clothes\_socks) \\
\texttt{hair} & 10 & (same as clothes\_socks) \\
\texttt{recent\_listen} & 13 & rock, jazz, disco, classical, funk... \\
\texttt{recent\_eat} & 10 & pizza, pasta, burrito, sushi, taco... \\
... & ... & ... \\ \bottomrule
\end{tabular}
\end{table}

\subsubsection{Puzzle Initialization}

A puzzle starts with initialization statements ($t=0$) that assign unique attribute values to each person: $\forall p \in P, \forall c \in A: S_0(p, c) \in V_c$ such that $\forall p_i, p_j \in P, i \neq j$, $\exists c \in A: S_0(p_i, c) \neq S_0(p_j, c)$.

\subsubsection{Statement Generation Process}

For each step $t$ from 1 to $N$, a statement is generated that changes the state of a  person. If it updates the PoI, the statement is called a \emph{needle}. An update for a non-PoI is called a \emph{hay}.

1. \textbf{Statement Type Selection}: Given $N$ and $\rho$, let $n_{\text{needle}}^t$ and $n_{\text{hay}}^t$ be the remaining numbers of needles and hays to satisfy the desired proportion $\rho$ in the complete puzzle. The probability of selecting a needle statement is then $\mathbb{P}(T_t = \text{needle}) = {n_{\text{needle}}^t}/{( N-t)}$. The total number of needle statements in the puzzle is calculated as $n_{\text{needle}}^0 = \max(1, \min(N, \text{round}(N \cdot \rho/100)))$.

2. \textbf{Reference Person Selection}: Given the selected statement type $T_t$, the algorithm selects the reference person $r_t$:  if $T_t= \text{needle} \implies r_t = p^*$ and if $T_t = \text{hay} \implies r_t \sim  \text{Uniform}(P \setminus \{p^*\})$.

3. \textbf{Statement Structure}: For each statement, CogniLoad samples a number of conditions $k_t \sim \text{Uniform}\{1, \ldots, d\}$, and a number of state updates $m_t \sim \text{Uniform}\{1, \ldots, d\}$ and uniformly sample attribute categories $C_t \subseteq A,\; |C_t| = k_t$ and state updates $U_t \subseteq A,\; |U_t| = m_t$.

4. \textbf{Condition and Update Value Specification}: For each category $c \in C_t$, the condition value is set by the reference person's current state: $v_{c,t} = S_{t-1}(r_t, c)$. For needles, these conditions target the PoI, while for hays the conditions can match multiple people. For update values, if $T_t = \text{needle} \implies u_{c,t} \sim \text{Uniform}(V_c)$ and if $T_t = \text{hay} \implies u_{c,t} \sim \text{Uniform}(V_c \setminus \{S_{t-1}(p^*, c)\})$.

5. \textbf{Logical Form}: The statement at step $t$ has the logical form:
  \begin{align*}
   \forall p \in P: \Big(\bigwedge_{c \in C_t} S_{t-1}(p, c) = v_{c,t}\Big) \Rightarrow \Big(\bigwedge_{c \in U_t} S_t(p, c) = u_{c,t}\Big).
   \end{align*}
   Attributes not mentioned in the update set remain unchanged $\forall p \in P, \forall c \in A \setminus U_t: S_t(p, c) = S_{t-1}(p, c)$. This is not specified in the prompt but implicitly assumed by the LLMs.

\subsubsection{Validation Constraints}

A sequence of validations verifies that the generated statement does not result in a state that prevents the generation of further needles and hays. If all validations pass, the statement is appended to the puzzle; otherwise a new statement is generated. 

\textit{For hay statements ($r_t \neq p^*$):} After the update, the state of affected non-PoIs must not become identical to PoI $\forall p \in P \setminus \{p^*\}$ such that $\forall c \in C_t: S_{t-1}(p, c) = v_{c,t}$, $\exists c \in A: S_t(p, c) \neq S_t(p^*, c)$ and the update must not affect the PoI $\exists c \in C_t: S_{t-1}(p^*, c) \neq v_{c,t}$.

\textit{For needle statements ($r_t = p^*$):} The update must not affect all non-PoI people $\exists p \in P \setminus \{p^*\}: \exists c \in C_t: S_{t-1}(p, c) \neq v_{c,t}$ and after the update not all non-PoIs have identical states as the PoI $\exists p \in P \setminus \{p^*\}: \exists c \in A: S_t(p, c) \neq S_t(p^*, c)$.

To prevent the distractors from becoming too trivial to track at lower difficulties, we require that a hay statement does not result in all non-PoIs become identical so the set $ P \setminus \{p^*\}$ must contain at least two persons with distinct attribute values. CogniLoad construction ensures that each hay statement $T_t = \text{hay}$   affects at least one non-PoI $\exists p \in P \setminus \{p^*\}: \forall c \in C_t: S_{t-1}(p, c) = v_{c,t}$.

\subsubsection{Final Question Generation}

After all $N$ statements have been generated, the puzzle concludes with a question about a random attribute of the PoI, sampled as a random category $c_q \sim \text{Uniform}(A)$. The correct answer to the puzzle is $S_N(p^*, c_q)$ obtained from the final state of the PoI.

\subsubsection{Evaluation metrics}
We evaluate each puzzle by exact-matching of the queried attribute value in the output of model $M$ with accommodating minor phrasing and common lexical variants. See Appendix \cref{supp-app:eval} for the specifics of the evaluation pipeline and an overview of the granular failure types that contextualize model specific error modes. The accuracy of model $M$ across the evaluation set is calculated as
$
\text{acc}(M)=
\frac{1}{|Z|}\sum_{z\in Z}
\mathbf{1}\bigl[\text{answer}_M(z)=S_N(p^*, c_q)\bigr]
$ where $S_N(p^*, c_q)$ represents the final state value of the queried attribute $c_q$ for the PoI $p^*$ after all $N$ statements have been processed.

\subsection{Tunable Parameters}
\label{subsec:hyperparameters}

To systematically probe long-context reasoning,  CogniLoad employs three independent parameters. These parameters are designed to operationalize distinct cognitive load dimensions as defined by CLT \citep{paas2003cognitive}, allowing the creation of puzzles with varying characteristics. Together, they define the load profile of a puzzle instance.

\paragraph{Intrinsic Difficulty $(d)$}
\label{subsubsec:difficulty_updated} for $d \in \{1,3,5,7,10\}$ controls multiple facets of puzzle complexity (see Table \ref{tab:hyperparameters}), directly manipulating ICL which according to CLT hinges on element interactivity \citep{halford1998processing}. Increasing $d$ increases ICL via: (i) combinatorial growth in state space ($\approx (d+1)^d$), (ii) increased interactivity between persons, attributes, and values, and (iii) increased rule complexity (up to $d$ conditions/updates per statement).

\paragraph{Task Length $(N)$} for $N \in \{20,50,100,250\}$ sets the total number of sequential state-update statements. While directly determining sequence length, $N$ serves as an operational proxy for conditions demanding GCL.  Increasing $N$, particularly with a large $d$ and $\rho$, compels deeper reasoning through more essential interacting elements \citep{sweller2010element}. Additionally, increasing $N$ necessitates the maintenance of a coherent (stateful) problem representation over a longer term with the construction of an efficient schema for it \citep{ericsson1995long}.

\begin{table}[t]
\centering
\caption{Key parameters controlling the puzzle generation.}
\label{tab:hyperparameters}
\begin{tabular}{lp{1.7cm}p{5.5cm}p{4cm}}
\toprule
Symbol & Name & Definition & Cognitive Load Affected \\
\midrule
$d$ & Intrinsic Difficulty &
Controls cardinality of people set $|P| = \max(d,2)$, attribute categories $|A| = d$, for each category $c \in A$ the cardinality of value domains $|V_c| = \max(d+1,3)$, and the distribution of conditions and updates per statement: $k,m \sim \text{Uniform}\{1,...,d\}$. 
& ICL: Element interactivity, state space/rule complexity. \\
\addlinespace
$N$ & Task Length &
Total number of sequential state transitions in the puzzle. &
GCL Proxy / Task Length: Demands sustained engagement with core elements. \\
\addlinespace
$\rho$ & Needle-to-hay Ratio &
Percentage of statements directly influencing the PoI (needles) versus distractor statements (hay) &
ECL: Distractor density challenges filtering, selective attention, and imposing load from processing non-essential elements. \\
\bottomrule
\end{tabular}
\end{table}

\paragraph{Needle-to-hay Ratio $(\rho)$}

for $\rho \in \{5,...,95\}$ sets the percentage of PoI-relevant (``needle'') versus distractor (``hay'') statements, directly manipulating ECL. ECL arises from processing non-essential elements \citep{chandler1991cognitive}. Decreasing $\rho$ increases ECL via increased distractor density which challenges filtering. Increasing $\rho$ controls ECL by focusing resources on relevant information. Critically, CogniLoad's ``hay'' statements are syntactically similar to ``needles'' and involve valid state updates for non-PoIs, imposing a more challenging ECL than easy to distinguish distractor text.

\section{Results}
\label{sec:results}

We have evaluated the performance of 13 open weights LLMs on 100 random CogniLoad puzzles per $(d, N, \rho)$ configuration resulting in 14'000 puzzle instances per LLM in total. In addition, we have evaluated the proprietary Gemini-2.5 and gpt-5\footnote{The gpt-5 family models were evaluated using the ``medium'' reasoning effort setting.} models and DeepSeek-R1-0528 on 10 CogniLoad puzzles per configuration (i.e., 1'400 puzzle instances). The maximum context length (input + output) was set to 32K tokens and LLMs run with their preset default decoding settings and system prompts. 

\cref{fig:results} shows mean accuracy across models as each load dimension varies with trends corroborated by our regression analysis (\cref{sec:logistic}).

\textbf{Intrinsic Difficulty (\(d\))} Performance declines monotonically with \(d\) for most models, although a few mid-tier models show small bumps at \(d=3\) (e.g., QwQ-32B: 0.62\(\rightarrow\)0.69; DS-Qwen-32B: 0.63\(\rightarrow\)0.66). Top models degrade only slightly from \(d=1\) to \(d=3\) (o3: 0.96\(\rightarrow\)0.93; gpt-5: 1.00\(\rightarrow\)0.97), while smaller or distilled models drop by 0.10-0.25 in the same range. By \(d=5\), 12 of 22 models fall below 50\% accuracy. Beyond \(d \geq 7\) the marginal decline flattens for the majority of models: the strongest models maintain their performance even at \(d=10\) (gpt-5: 0.82; o3: 0.80), and the weakest ones approach 0.10-0.15.

\textbf{Task Length ($N$)} This parameter remains the dominant stressor. Most models exhibit their steepest decline between \(N=20\) and \(N=50\) (e.g., DS-Llama-70B: 0.89\(\rightarrow\)0.66; Qwen3-8B: 0.71\(\rightarrow\)0.40), while the best performing ones show relative resilience (gpt-5: 1.00\(\rightarrow\)0.98; o3: 0.99\(\rightarrow\)0.98). Accuracy declines with longer sequences as at \(N=100\), several models roughly halve their \(N=20\) accuracy (e.g., DS-Llama-70B: 0.48; Qwen3-32B: 0.38) and at \(N=250\) only two models show above 50\% accuracy (i.e., gpt-5 at 0.76 and o3 at 0.68) while the majority ones perform  0.20-0.30 accuracy.

\textbf{Extraneous Load / Needle-to-hay Ratio (\(\rho\))} A characteristic U-shaped response is typical with performance usually reaches as low as \(\rho\in[25,50]\)\% and recovers as \(\rho\) increases. Recovered performance equals or exceeds that of small-\(\rho\) baseline in several cases (DS-Llama-70B: 0.61\(\rightarrow\)0.64; gpt-5-mini: 0.84\(\rightarrow\)0.86; gemini-2.5-flash-lite: 0.38\(\rightarrow\)0.53). The strongest models show smooth variations (gpt-5: 0.97\(\rightarrow\)0.89\(\rightarrow\)0.91; o3: 0.93\(\rightarrow\)0.89\(\rightarrow\)0.88), indicating marginal sensitivity to distraction, while some models recover only partially or not at all (Phi-4-reasoning-plus: 0.59\(\rightarrow\)0.45; EXAONE-Deep-32B: 0.47\(\rightarrow\)0.32).

\subsection{Load-sensitivity Regression}
\label{sec:logistic}
To quantify model-specific sensitivities of the accuracy to load dimensions and derive interpretable capacity thresholds for each model, we employ a regression-based approach that allows us to isolate the impact of each type of cognitive load (see  \cref{tab:coeff-quad}).

\begin{table}[t]
\centering
\small
\caption{Per-model quadratic‐$\rho$ GLM estimates with Wald $z$ statistic for p-values alongside derived 50\% load‐capacity thresholds (see \cref{sec:capacity-points}). The value $--$ for NT$_{50}$ indicates that no real root exists in $[0,1]$. ``DS'' abbreviates ``DeepSeek'' in the model names. $^{***}$$p{<}0.001$, $^{**}$$p{<}0.01$, $^{*}$$p{<}0.05$}
\setlength{\tabcolsep}{4pt}
\begin{tabular}{lrrrrrrrr}
\toprule
Model & $\beta_{0}$ & $\beta_{d}$ & $\beta_{N}$ & $\beta_{\rho}$ & $\beta_{\rho^2}$ & ECL$_{50}$ & NT$_{50}$ & ID$_{50}$ \\
\midrule
gemini-2.5-pro & $22.51^{***}$ & $-0.41^{***}$ & $-9.15^{***}$ & $-1.67$ & $1.76$ & $153.3$ & $--$ & $12.74$  \\
gemini-2.5-flash & $18.14^{***}$ & $-0.44^{***}$ & $-7.56^{***}$ & $-1.79$ & $2.16$ & $111.5$ & $--$ & $8.56$  \\
gemini-2.5-flash-lite & $3.19^{***}$ & $-0.30^{***}$ & $-1.22^{***}$ & $-2.88^{**}$ & $3.82^{***}$ & $8.8$ & $0.93$ & $1.53$  \\

gpt-5-2025-08-07 & $17.34^{***}$ & $-0.39^{***}$ & $-5.11^{***}$ & $-7.04^{***}$ & $5.62^{***}$ & $382.8$ & $--$ & $14.78$  \\
gpt-5-mini-2025-08-07 & $11.09^{***}$ & $-0.22^{***}$ & $-3.96^{***}$ & $-4.87^{***}$ & $5.10^{***}$ & $164.1$ & $--$ & $11.72$  \\
gpt-5-nano-2025-08-07 & $6.50^{***}$ & $-0.31^{***}$ & $-2.43^{***}$ & $-4.30^{***}$ & $4.12^{***}$ & $35.7$ & $0.94$ & $2.87$  \\
o3-2025-04-16 & $12.83^{***}$ & $-0.22^{***}$ & $-4.26^{***}$ & $-2.72$ & $2.07$ & $356.9$ & $--$ & $19.07$  \\
o4-mini-2025-04-16 & $13.00^{***}$ & $-0.23^{***}$ & $-4.89^{***}$ & $-5.99^{***}$ & $6.37^{***}$ & $132.1$ & $--$ & $10.86$  \\
DS-R1-0528 & $13.70^{***}$ & $-0.39^{***}$ & $-5.28^{***}$ & $-4.13^{***}$ & $4.19^{***}$ & $104.6$ & $--$ & $7.51$  \\
DS-Llama-70B & $8.36^{***}$ & $-0.30^{***}$ & $-3.28^{***}$ & $-3.50^{***}$ & $3.92^{***}$ & $69.8$ & $0.53$ & $5.14$  \\
DS-Qwen-32B & $5.15^{***}$ & $-0.19^{***}$ & $-2.12^{***}$ & $-2.07^{***}$ & $2.29^{***}$ & $54.3$ & $0.78$ & $3.95$  \\
DS-Qwen-7B & $1.74^{***}$ & $-0.23^{***}$ & $-0.96^{***}$ & $-0.45$ & $0.58^{*}$ & $2.9$ & $--$ & $-0.53$  \\
DS-Qwen-1.5B & $-0.35^{**}$ & $-0.20^{***}$ & $-0.33^{***}$ & $0.47$ & $-0.14$ & $0.0$ & $--$ & $-3.95$  \\
Phi-4-reasoning-plus & $9.52^{***}$ & $-0.45^{***}$ & $-3.58^{***}$ & $-4.21^{***}$ & $3.41^{***}$ & $45.7$ & $0.16$ & $3.68$  \\
Phi-4-reasoning & $9.11^{***}$ & $-0.39^{***}$ & $-3.35^{***}$ & $-4.62^{***}$ & $3.99^{***}$ & $52.3$ & $0.92$ & $4.08$  \\
Phi-4-mini-reasoning & $1.70^{***}$ & $-0.24^{***}$ & $-0.81^{***}$ & $-1.40^{***}$ & $1.45^{***}$ & $1.3$ & $--$ & $-0.55$  \\
EXAONE-Deep-32B & $4.09^{***}$ & $-0.26^{***}$ & $-1.55^{***}$ & $-3.16^{***}$ & $2.45^{***}$ & $14.1$ & $--$ & $1.0$  \\
QwQ-32B & $5.68^{***}$ & $-0.21^{***}$ & $-2.08^{***}$ & $-3.70^{***}$ & $3.07^{***}$ & $48.0$ & $0.95$ & $3.56$  \\
Qwen3-32B & $7.22^{***}$ & $-0.29^{***}$ & $-2.75^{***}$ & $-3.21^{***}$ & $2.75^{***}$ & $53.9$ & $0.94$ & $4.1$  \\
Qwen3-30B-A3B & $5.83^{***}$ & $-0.30^{***}$ & $-2.25^{***}$ & $-2.96^{***}$ & $2.87^{***}$ & $36.7$ & $0.99$ & $3.05$  \\
Qwen3-8B & $5.40^{***}$ & $-0.26^{***}$ & $-2.19^{***}$ & $-2.87^{***}$ & $2.66^{***}$ & $30.8$ & $--$ & $2.2$  \\
Qwen3-1.7B & $0.62^{***}$ & $-0.17^{***}$ & $-0.46^{***}$ & $-1.53^{***}$ & $1.24^{***}$ & $0.0$ & $--$ & $-4.07$  \\

\bottomrule
\end{tabular}

\label{tab:coeff-quad}
\end{table}

\subsubsection{Regression Model Specification}

We model the performance of LLMs using a binomial generalized linear model (GLM) with a logit link function:
\[
\Pr\!\bigl(Y{=}1\bigr) = \sigma\!\bigl(
  \beta_{0}
+ \beta_{d}\,d
+ \beta_{N}\,\log_{10}N
+ \beta_{\rho}\,\rho
+ \beta_{\rho^2}\,\rho^2
\bigr),
\]
where the binary outcome $Y$ represents exact-match accuracy ($Y=1$, when the model solves the puzzle correctly), $\sigma(\cdot)$ is the inverse logit function, and the coefficients $\beta_{d}$, $\beta_{N}$ and $\beta_{\rho}$  quantify sensitivity to intrinsic difficulty (ICL), task length (GCL), and distractor ratios (ECL), respectively. The inclusion of a quadratic term for $\rho$ is motivated by the characteristic U-shape observed in the third panel of~\cref{fig:results} and based on an improved Akaike Information Criterion (AIC) value for 18 out of the 22 fitted models when included (see Appendix \cref{supp-app:aic}). Since $N$ ranges up to 250, we apply $\log_{10}$ to keep it at a similar scale as the other parameters of the regression.

\subsubsection{Significance of Main Effects}

In all models, $\beta_d$ and $\beta_N$ are significant and highly negative, confirming performance degradation with increased ICL and GCL. The quadratic term for $\rho$ is also significant (except for two models) confirming the U-shaped response for most models: models typically perform worst at intermediate $\rho$ values and recover as $\rho$ approaches either extreme. Five models exhibit statistically insignificant coefficients for $\rho$ terms, reflecting poor baseline performance for the smallest models and indifference to distraction for strong models (i.e., o3, gemini-2.5-pro, gemini-2.5-flash).

\subsubsection{Capacity Points at 50\% Accuracy}
\label{sec:capacity-points}
The GLM coefficients (\cref{tab:coeff-quad}) allow us to derive interpretable capacity thresholds. These represent the point at which a model's accuracy is predicted to drop to 50\% when varying a single load parameter, while holding other load parameters at their estimated mean values:

\textbf{ECL$_{50}$} (Effective Context Length): Maximum number of statements a model can process while maintaining 50\% accuracy. Large ECL$_{50}$ values indicate superior context handling.

\textbf{NT$_{50}$} (Needle-to-hay Threshold): Minimum proportion of relevant information required to maintain 50\% accuracy. Crucially, \textit{small} values indicate greater robustness to distractors. If the estimated NT$_{50}$ is missing, then the model accuracy is not expected to cross the 50\% threshold for any value  $0 \le \rho \le 1$, under mean conditions for $d$ and $N$.

\textbf{ID$_{50}$} (Intrinsic Difficulty): It is the maximum intrinsic complexity (number of interacting entities/attributes) that a model can handle while maintaining 50\% accuracy. Negative values indicate failure to reach 50\% accuracy even at the lowest difficulty setting under mean conditions for $N$ and $\rho$.

Mathematically, these thresholds are derived by setting the logit in the GLM equation to zero (for $\Pr(y=1)=0.5$) and solving for the parameter of interest, e.g.:
\[
\small
\mathrm{ECL}_{50} = 10^{-({\beta_{0}+\beta_{d}\bar{d}+\beta_{\rho}\bar{\rho}+\beta_{\rho^2}\bar{\rho}^2})/{\beta_{N}}}; \quad
\mathrm{ID}_{50} = -(\beta_{0}+\beta_{N}\overline{\log_{10} N}+\beta_{\rho}\bar{\rho}+\beta_{\rho^2}\bar{\rho}^2)/{\beta_{d}}.
\]
For $\mathrm{NT}_{50}$, we solve the quadratic equation $\beta_{0}+\beta_{d}\bar{d}+\beta_{N}\overline{\log_{10} N}+\beta_{\rho}\rho+\beta_{\rho^2}\rho^2 = 0$ for $\rho$.

\subsubsection{Model Capacity}
The regression analysis and estimated capacity thresholds (Table~\ref{tab:coeff-quad}) reveal clear variations among models that can be grouped into three classes:

\textit{Frontier/High-capacity Models:} gpt-5 and o3 lead by a wide margin (ECL${50}$ > 300), followed by gemini-2.5-pro, gpt-5-mini, o4-mini, and DS-R1-0528. The high baseline performance of gemini-2.5-pro ($\beta_0$ = 22.5) together with the large $\beta_N$ of -9.15 is consistent with the uniquely large amount long context errors as $N$ increases (as illustrated in the Appendix \cref{supp-app:err_dist}).

\textit{Mid-capacity Models:} DS-Llama-70B , Qwen3-32B, DS-Qwen-32B, QwQ-32B, Phi-4-reasoning, Phi-4-reasoning-plus, Qwen3-30B-A3B, gpt-5-nano-2025-08-07, and Qwen3-8B form a broad middle tier with good performance at moderate $N$ and $d$ values.

\textit{Low-capacity Models:} DS-Qwen-7B, Phi-4-mini-reasoning, DS-Qwen-1.5B, and Qwen3-1.7B exhibit minimal effective context handling capacity failing to reach 50\% accuracy even under mean context/distractor conditions, deteriorating rapidly under slightly increasing load. 

\subsubsection{Differential Sensitivity to Load Dimensions}
The estimated coefficients further reveal distinct sensitivity profiles:

\textit{Sensitivity to context length} ($\beta_N$): Universally negative and highly significant, larger models often show greater relative degradation compared to their higher baselines. Yet large ECL$_{50}$ values for frontier models arise from the combined effect of \(\beta_{0}\), \(\beta_N\) indicating comparisons are best made via ECL$_{50}\) and not \(\beta_N\) in isolation.

\textit{Sensitivity to intrinsic difficulty} ($\beta_d$): 
Negative across models with a narrow range, it suggests a more uniform effect. Despite some steep \(\beta_d\) values, high baselines (e.g., gemini-2.5-flash) yield large ID$_{50}$ values unlike smaller models with similar \(\beta_d\) (e.g., Phi-4-reasoning-plus). 

\textit{Sensitivity to information relevance} ($\beta_{\rho}$ and $\beta_{\rho^2}$): Confirms the U-shaped response, but NT$_{50}$ values reveal nuanced distractor robustness variations masked by aggregate scores (e.g., DS-Llama-70B vs. Qwen3-32B). For frontier models, the absence of NT$_{50}$ indicates achieving above 50\% accuracy while for weak models the same absence indicates remaining below 50\% accuracy.

\subsection{Failure modes across models, length, and difficulty}
We analyze error categories from the evaluation pipeline (see Appendix \cref{supp-app:eval}) to identify failure modes and provide the complete distributions and per-model breakdowns in the Appendix \cref{supp-app:err_dist}.

\textbf{State-tracking mistakes dominate under load.} Across models, the most common non-context failure is wrong final attribution in the last valid PoI sentence (valid-logic), consistent with mis-tracking sequential updates rather than formatting issues. For example, at $N$=250, Qwen3-32B has 2'541 valid-logic cases, DS-Llama-70B 2'465, and QwQ-32B 2'092. These logic errors also increase monotonically with $d$ for nearly all models.

\textbf{Long-context budget overflows are a prominent, model-specific failure at extreme $N$.} The max-context errors grow sharply with $N$ for some models: gemini-2.5-flash (280 errors in 350 samples at $N=250$) or gemini-2.5-pro (268/350). OpenAI models also make these errors at $N=250$ but at much lower levels (gpt-5: 32/350; o3: 24/348). The high error counts for the Gemini models indicate relatively poor token efficiency when reasoning. 

\textbf{Instruction-following drift emerges under higher $N$ and $d$, mainly in smaller models.} While poi-logic stays near zero for most models, last-logic increases notably for compact models (e.g., Phi-4-mini-reasoning: 400 last-logic at $N=250$); DS-Qwen-7B: 116), indicating that under load, models often fail to answer in the instructed format.

\textbf{"Other" failures rise with sequence length in small and mid-tier models.} At $N=250$, DS-Qwen-1.5B has 605 “other” cases (often claiming the puzzle is unsolvable) and DS-Qwen-7B 461 indicating a shift from precise (but wrong) answers to non-answers as load grows.

\section{Discussion}

CogniLoad, by operationalizing CLT, enables a multi-dimensional evaluation of LLM reasoning, revealing nuanced failure patterns obscured by single-dimensional benchmarks. Our empirical results (\cref{sec:results}) offer several key insights: task length ($N$) emerges as a dominant determinant, suggesting challenges in sustained, germane-like processing for long, intrinsically demanding tasks; models exhibit distinct sensitivities to intrinsic difficulty ($d$) versus extraneous load ($\rho$), with the latter surprisingly showing U-shaped performance curves, indicating particular difficulties with intermediate distractor densities; and estimated capacity thresholds provide concise ``cognitive fingerprints'' for diagnostic LLM evaluation. The limitations of our study are summarized as follows:

\textbf{Nuances of the CLT-LLM Analogy} While CLT provides a powerful analogous framework, it is crucial to acknowledge that ``cognitive load'' in LLMs manifests as computational constraints (e.g., attention saturation, representational bottlenecks) rather than biological working memory limitations. Our operationalization of $N$ as a proxy for conditions demanding GCL, for example, is an abstraction. Future research should aim to bridge CLT concepts with direct, mechanistic measures of the underlying  computational processes in LLMs to refine this analogy.

\textbf{Scope of Reasoning and Generalizability} CogniLoad focuses on sequential and pure deductive reasoning without requiring domain knowledge. While this reasoning type is fundamental to various subject areas (e.g., code, math), it is distinct from alternative reasoning paradigms like inductive, abductive, or analogical reasoning. Extending the CLT-grounded multi-dimensional evaluation to other reasoning types and evaluating it in other languages is a promising next step.

\textbf{Beyond Accuracy and Main Effects} The current evaluation relies on exact-match accuracy. Future iterations could incorporate richer metrics (e.g., step-wise reasoning fidelity, solution coherence, uncertainty of solutions) and systematically investigate interactions among $d, N, \rho$, which CogniLoad's factorial design supports. Reinforcement learning on verifiable rewards \citep{guo2025deepseek} presents a promising application of CogniLoad in LLM training, as its generated metadata enables precise verification of reasoning steps despite limited data of this kind.

Despite these considerations, by decomposing the ``task difficulty'' into principled, controllable dimensions inspired by CLT, CogniLoad provides a more insightful perspective than single-dimensional benchmarks. It allows a more differentiated understanding of LLM reasoning capabilities and limitations, paving the way for more targeted development of robust and generalizable AI systems. 

\section{Conclusion}
We introduced \textbf{CogniLoad}, a novel synthetic benchmark grounded in CLT for multi-dimensional evaluation of LLM long-context reasoning. By independently controlling parameters for intrinsic cognitive load ($d$), extraneous cognitive load ($\rho$), and task length ($N$), CogniLoad offers unprecedented diagnostic precision. Our evaluations revealed task length as a dominant performance constraint and uncovered unique ``cognitive fingerprints'' of LLM sensitivities to different load types, providing actionable insights beyond single-dimensional benchmarks. CogniLoad offers a reproducible, scalable, and theoretically-grounded tool to systematically dissect LLM reasoning limitations and guide the development of more capable and robust AI systems. While human and artificial cognition are mechanistically distinct, applying frameworks such as CLT to AI evaluation can provide valuable perspectives for understanding and characterizing their operational differences and capabilities.

\section{LLM Disclosure}
LLMs were used only in the early stages of writing the paper to refine phrasing and correct grammar. During coding they were used to update the chart formatting code, and to translate the manual implementation of the generation algorithm in Elixir to python for reproducibility.

\section{Reproducibility}
The code for generating CogniLoad puzzles according to the algorithm described in this paper is provided at: \url{https://anonymous.4open.science/r/cogniload-292B/}. The dataset of the puzzles on which the LLMs were evaluated for the results presented in this paper is provided on HuggingFace:
\url{https://huggingface.co/datasets/cogniloadteam/cogniload}

\section{Author Contributions}
All authors contributed in editing the manuscript towards the final version. D.K conceived the idea, designed and implemented the generation algorithm, prepared the dataset, scored the LLMs, analysed the empirical results, designed and implemented the evaluation pipeline, and wrote the first draft of the manuscript. All authors refined the benchmark design. B.R. compiled the first draft of the literature overview and acquired the computational resources for the evaluation. 

\section{Acknowledgements and Support}

This work was supported by the Research Council of Norway through its Centres of Excellence scheme, Integreat - Norwegian Centre for knowledge-driven machine learning, project number 332645. 

Ali Ramezani-Kebrya was supported by the Research Council of Norway through FRIPRO Grant under project number 356103, its Centres of Excellence scheme, Integreat - Norwegian Centre for knowledge-driven machine learning under project number 332645, and its Centre for Research-based Innovation funding scheme (Visual Intelligence under grant no. 309439).

We acknowledge NRIS Norway for awarding this project access to the LUMI supercomputer, owned by the EuroHPC Joint Undertaking, hosted by CSC (Finland) and the LUMI consortium through Sigma2, project number nn12027k.

\bibliographystyle{abbrvnat}
\bibliography{bibliography}

\end{document}

\maketitle

\noindent The supplementary material is organized as follows: 
\begin{itemize}
    \item ~\cref{app:rho} investigates $\rho$ relative to $N$ and $d$.
    \item ~\cref{app:error-bars} charts model performance reported with error-bars.
    \item ~\cref{app:eval} discusses the evaluation, scoring pipeline, and error taxonomy.
    \item ~\cref{app:err_dist} reports the distribution of error types across models and different levels of $d$ and $N$.
    \item ~\cref{app:aic} compares the AIC values of linear vs quadratic $\rho$ in the GLM regression.
    \item ~\cref{app:ontology} details the complete attribute ontology.
 \end{itemize}

\section{Investigating $\rho$ relative to $N$ and $d$}
\label{app:rho}

\begin{figure}[h!]
    \centering
    \includegraphics[width=\textwidth]{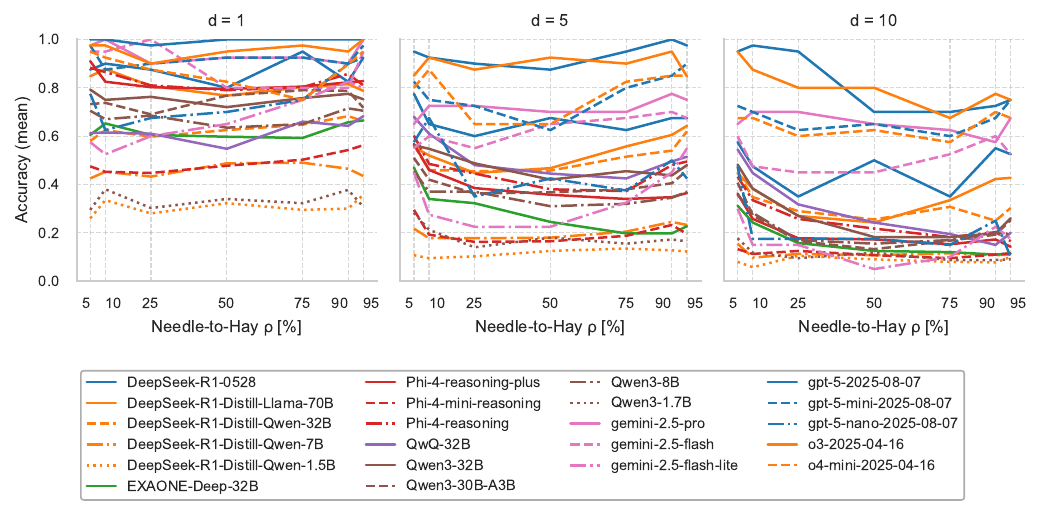}
    \caption{Distractor ratio $\rho$ relative to $d$ for $d \in \{1,5,10\}$}
    \label{fig:rho_diff}
\end{figure}

\begin{figure}[h!]
    \centering
    \includegraphics[width=\textwidth]{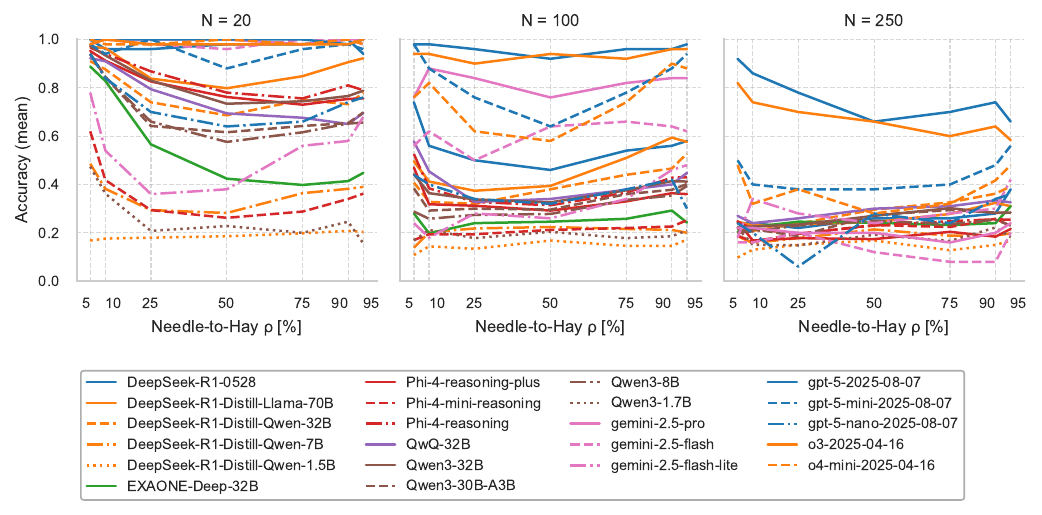}
    \caption{Distractor ratio $\rho$ relative to $N$ for $N \in \{20,100,250\}$}
    \label{fig:rho_N}
\end{figure}

In Figure 1 of the paper we discover a characteristic U-shape of the performance of LLMs on CogniLoad relative to $\rho$ (i.e., the ratio of distractors to essential elements). The following figures investigate this U-shape at different levels of difficulty (Figure \ref{fig:rho_diff}) and statement length (Figure \ref{fig:rho_N}).

Figure \ref{fig:rho_diff} plots mean accuracy as a function of the needle‑to‑hay ratio \( \rho \) (higher \( \rho \) means more signal per distractor) under three difficulty settings \( d = 1, 5, 10 \). For the easiest tasks (\( d = 1 \), left) accuracies are high and largely flat; most models show at most a shallow U‑shape, with a mild dip around \( \rho = 25\,\% \)–\( 50\,\% \) and a small recovery by \( \rho \ge 75\,\% \). At intermediate difficulty (\( d = 5 \), centre) the trough widens: many curves drop sharply from \( \rho = 10\,\% \) to \( 25\,\% \) and bottom out across \( \rho \approx 25\,\% \)–\( 50\,\% \) before rebounding at high \( \rho \). A notable exception is EXAONE‑Deep‑32B, which trends downward as \( \rho \) increases for \( d \ge 5 \). Under the hardest condition (\( d = 10 \), right) capability tiers separate. Frontier models remain robust and still recover at high \( \rho \)—averaged over \( \rho \), GPT‑5 \(0,821\), O3 \(0,804\), GPT‑5‑mini \(0,675\), O4‑mini \(0,646\), and Gemini‑2.5‑Pro \(0,654\)—whereas mid‑size and small models stay low despite a late uptick (e.g., DeepSeek‑R1‑0528 \(0,475\); DeepSeek‑R1‑Distill‑Llama‑70B \(0,361\); QwQ‑32B \(0,300\); Qwen3‑32B \(0,280\); Phi‑4‑reasoning \(0,256\); Qwen3‑1{,}7B \(0,110\)). Overall, the panels show that (i) accuracy shifts downward and the U‑shape becomes more pronounced as \( d \) increases; (ii) recovery at high \( \rho \) is strongly model‑dependent and favours larger, reasoning‑tuned systems; and (iii) the moderate‑\( \rho \) regime \( (\rho \approx 25\,\% \text{–} 50\,\%) \) remains the most adversarial, where balancing signal and noise is hardest.

Figure \ref{fig:rho_N} shows mean accuracy as a function of the needle‑to‑hay ratio \( \rho \) for \(N = 20\) (left), \(N = 100\) (centre), and \(N = 250\) (right). At \(N = 20\) many systems still display a clear U‑shape: accuracy begins near \(1,0\), dips at medium clutter around \( \rho \approx 25\)–\(50\,\% \), and then rebounds as needles dominate. As the context grows, this dependency compresses and then largely disappears. At \(N = 100\) the strongest models are nearly \( \rho \)-invariant (e.g., curves remain between about \(0,93\) and \(0,97\)), while mid‑tier models retain only a shallow trough. By \(N = 250\) most curves are flat or gently increasing with \( \rho \) at much lower levels; only the frontier models exhibit a modest late uptick.

Taken together, the two figures show that difficulty \(d\) and length \(N\) both depress accuracy but modulate its dependence on \( \rho \) in different ways. Increasing \(d\) accentuates the classic U‑shape (i.e., many models drop from \( \rho = 10\,\% \) into the mid‑range \( \rho \approx 25\,\% \)–\( 50\,\% \) and then recover as needles dominate) whereas increasing \(N\) shifts the whole curve downward and progressively flattens it. These stresses reveal a clear stratification: frontier models remain comparatively robust (retaining some high‑\( \rho \) gains and partial \( \rho \)‑invariance), a mid‑tier exhibits a pronounced but narrowing mid‑\( \rho \) trough as \(N\) grows, and small models settle at low, nearly \( \rho \)‑insensitive performance. In short, the mid‑\( \rho \) band \(25\)–\(50\,\% \) is the most adversarial at fixed \(N\) and \(d\), but at long contexts the primary failure mode becomes \(N\) itself rather than \( \rho \).

\section{Model performance with Error-Bars}
\label{app:error-bars}
Due the small size and the large amount of models we compare in Figure 1 of the main paper we omit error-bars to improve legibility. In this section we plot larger versions of the charts in the panel with vertical error bars representing 90\% confidence intervals for the mean accuracy of each model–condition pair.

We compute these confidence intervals with the Wilson score method for a binomial proportion.  Specifically, for a given point with \(k\) correct answers out of \(n\) trials we set \(\hat p = k/n\) and use the 90 \% standard-normal quantile \(z = 1.644853627\) to obtain
\[
\bigl[l,\;u\bigr] \;=\;
\frac{%
\hat p + \tfrac{z^{2}}{2n}
\;\pm\;
z\!\sqrt{\dfrac{\hat p\,(1-\hat p) + z^{2}/(4n)}{n}}
}%
{1 + \tfrac{z^{2}}{n}}.
\]

The resulting interval \([l,u]\) marks the range that would contain the true underlying accuracy in 90\% of repeated experiments with the same sample size.  Wider bars correspond to greater sampling uncertainty, whereas tighter bars indicate more stable estimates.

The narrow error bars we observe substantiate the discussion of the results in the main paper and highlights a  significant difference between the curves at 90\% significance and the presence of the characteristic U-shape. 

\begin{figure}[h!]
    \centering
    \includegraphics[width=\textwidth]{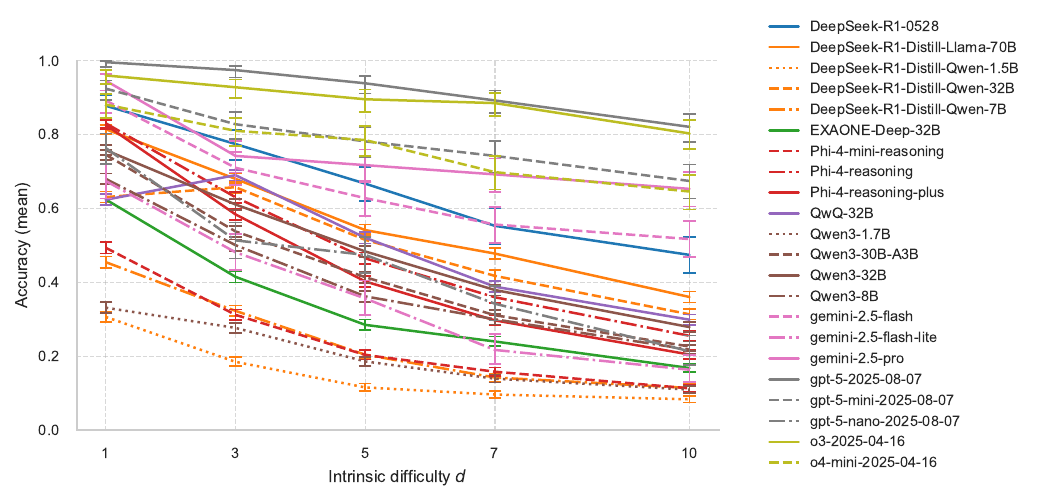}
    \caption{Accuracy relative to difficulty $d$ for $d \in \{1,3,5,7,10\}$ on the X-Axis}
    \label{fig:panel_d}
\end{figure}

\begin{figure}[h!]
    \centering
    \includegraphics[width=\textwidth]{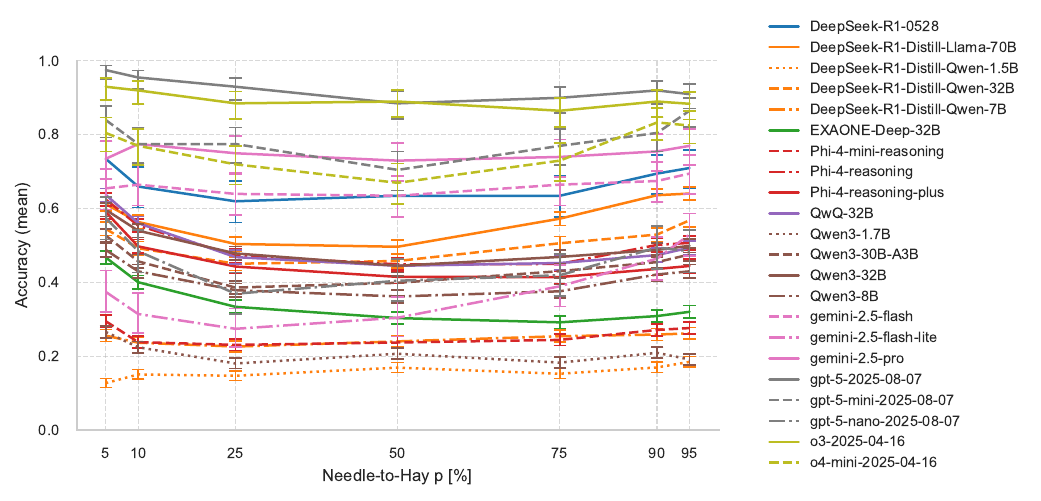}
    \caption{Accuracy relative to distractor-ratio $\rho$ for $\rho \in \{5,10,25,50,75,90,95\}$ on the X-Axis}
    \label{fig:panel_r}
\end{figure}

\begin{figure}[h!]
    \centering
    \includegraphics[width=\textwidth]{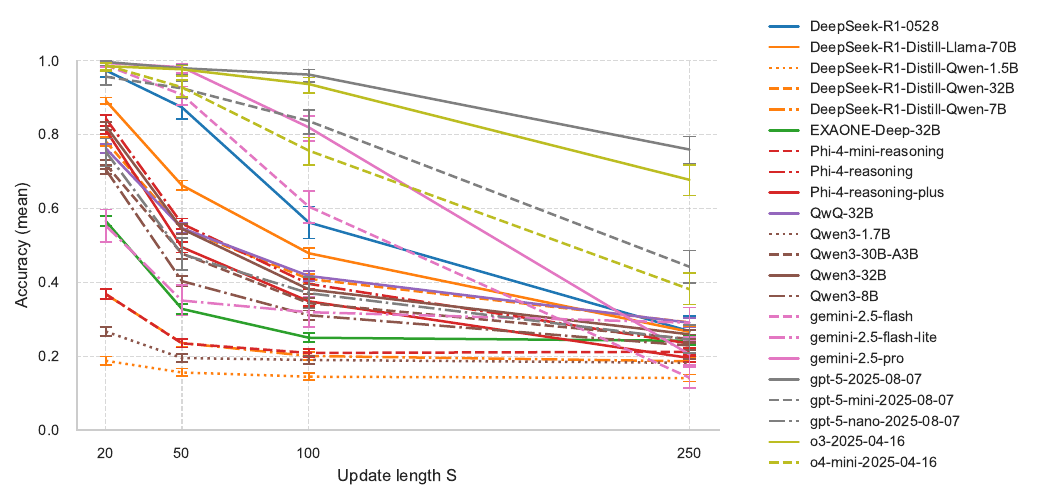}
    \caption{Accuracy relative to total statement length $N$ for $N \in \{20,50,100,250\}$ on the X-Axis}
    \label{fig:panel_N}
\end{figure}

\section{Evaluation, scoring pipeline, and error taxonomy}
\label{app:eval}
This section fully details the evaluation and scoring pipeline functionally in text and in pseudo code.

\subsection{Inputs, files, and naming}

The scorer consumes three artefacts per puzzle instance \(z\):
\begin{enumerate}
    \item Model output of the reasoning trace + response (plain text).
    \item Puzzle text (for the question): The last line is the natural-language question.
    \item Puzzle metadata (JSON state): providing the person-of-interest (PoI), gold label, needles, difficulty, statement count, and category domains.
\end{enumerate}

\subsection{Category mapping and qualifiers}

The target category is resolved from the last line question using deterministic templates:
\begin{itemize}
    \item \(\texttt{startswith("Where is")}\) \(\to\) \texttt{location}
    \item \(\texttt{startswith("What color shirt")}\) \(\to\) \texttt{clothes\_shirt}
    \item \(\texttt{startswith("What color pant")}\) \(\to\) \texttt{clothes\_pant}
    \item \(\texttt{startswith("What color hat")}\) \(\to\) \texttt{clothes\_hat}
    \item \(\texttt{startswith("What color of socks")}\) \(\to\) \texttt{clothes\_socks}
    \item \(\texttt{startswith("What color of gloves")}\) \(\to\) \texttt{clothes\_gloves}
    \item \(\texttt{startswith("What color of underwear")}\) \(\to\) \texttt{clothes\_underwear}
    \item \(\texttt{startswith("What is the final hair color")}\) \(\to\) \texttt{hair}
    \item \(\texttt{endswith("most recently eat?")}\) \(\to\) \texttt{recent\_eat}
    \item \(\texttt{endswith("recently watch?")}\) \(\to\) \texttt{recent\_watch}
    \item \(\texttt{endswith("recently listen to?")}\) \(\to\) \texttt{recent\_listen}
    \item \(\texttt{endswith("recently read?")}\) \(\to\) \texttt{recent\_read}
\end{itemize}

Qualifiers (used to find a ``valid PoI sentence''):
\begin{itemize}
    \item \texttt{location}: \{at, located, in\}
    \item \texttt{clothes\_shirt}: \{shirt, wear\}; \texttt{clothes\_pant}: \{pant, wear\}; \texttt{clothes\_hat}: \{hat, wear\}; \texttt{clothes\_gloves}: \{glove, wear\}; \texttt{clothes\_socks}: \{sock, wear\}; \texttt{clothes\_underwear}: \{underwear, wear\}
    \item \texttt{hair}: \{hair\}
    \item \texttt{recent\_eat}: \{eat, ate\}; \texttt{recent\_watch}: \{watch, watched, movie\}; \texttt{recent\_listen}: \{listen, listened, music\}; \texttt{recent\_read}: \{read, book\}
\end{itemize}

\noindent  The PoI name and the domain values for the target category are read from the puzzle metadata file.

\subsection{Normalization, sentence windows, and context budget}

\noindent Normalization:
\begin{itemize}
    \item Lowercase the full model output; split by newlines.
    \item If the last line starts with ``('' and ends with ``)'', drop that line because it's usually just a clarification that follows the final answer.
    \item Drop empty lines; if no lines remain, keep a single empty line.
\end{itemize}

\noindent Sentence windows:
\begin{enumerate}
    \item \textbf{valid\_poi\_last\_sentence}: last line containing PoI and any category qualifier; take \texttt{TakeLastSentence}.
    \item \textbf{poi\_last\_sentence}: last line containing PoI; take \texttt{TakeLastSentence} of that line.
    \item \textbf{last\_sentence}: apply \texttt{TakeLastSentence} to the last remaining line; see below.
\end{enumerate}

\noindent TakeLastSentence(\(s\)):
let \(p = \texttt{split}(s, \text{"."})\). If \(\texttt{len}(p) \ge 2\), return \(p[-2]\); else return \(s\).

\noindent Context limit failure:
we mark \texttt{wrong\_max\_context} if any of the following holds:
\[
\texttt{prompt\_tokens} + \texttt{response\_tokens} + 20 \ge 32'768, \quad \text{or} \quad \text{last\_sentence} = \text{""}, \quad \text{or} \quad \text{empty output}.
\]
The 20 additional tokens added to the \texttt{response\_tokens} is necessary because when models usually produce incomplete responses a few tokens before the maximum-output-token limit they were prompted to decode with.

\subsection{Matching, synonyms, and alternatives}

\noindent Since some models respond with the answer attribute starting with a special character instead of an empty space (like a word) the containment accepts the following alternative word boundries (code \texttt{CheckAnswer(term, sentence)} returns true if any holds):
\begin{itemize}
    \item \(\texttt{sentence.startswith(term)}\) or \(\texttt{sentence == term}\)
    \item substring appears prefixed by one of: space, ``['', ``"'', ``*'', ``\_ '', ``\{ '', ``(''
\end{itemize}

\noindent Since some attributes in the attribute taxonomy could be spelled alternatively, we relax 
Synonym whitelist for them in the evaluation (code \texttt{CheckCorrectAnswer(gold, sentence)}):
\begin{itemize}
    \item \texttt{reaggea}: accept \texttt{reaggea} or \texttt{reggae}
    \item \texttt{sci-fi}: accept \texttt{sci-fi}, \texttt{science fiction}, \texttt{science-fiction}
    \item \texttt{camp}: accept \texttt{camp}, \texttt{campground}
    \item \texttt{potatoes}: accept \texttt{potatoes}, \texttt{potato}
    \item \texttt{market}: accept \texttt{market}, \texttt{marketplace}
    \item \texttt{livingroom}: accept \texttt{livingroom}, \texttt{living room}
    \item otherwise: accept exactly \texttt{gold} by boundary-tolerant containment
\end{itemize}

\noindent We further ensure no alternative value is present in the last sentence and therefore  compute (\texttt{AltFlag(sentence, alts, gold)}) as:
\begin{enumerate}
    \item Let \(s=\) lowercase(sentence), \(g=\) lowercase(gold), and \(A=\) lowercase alternatives \(=\) domain values \(\setminus\{g\}\).
    \item Define \(\texttt{present}(t) = \texttt{CheckCorrectAnswer}(t, s)\).
    \item If no alternative is present, return false.
    \item If both gold and at least one alternative are present, keep the alternative flag only if some alternative’s character span encloses the gold’s span in \(s\) (``spans'' test); otherwise clear the flag. This is done to accommodate attribute values like 'non-fiction' which contain alternative values like 'fiction'.
    \item Since sometimes models respond by mentioning also previous value of the attribute (eg a sentence like ''\texttt{... the final step explicitly converts gray to green, resulting in **green**}'') we do not set the flag if the gold flag is the last mentioned attribute in the sentence.
\end{enumerate}

\subsection{Buckets, precedence, and accuracy}

Let:
\begin{align*}
c_{\mathrm{valid}} &= \texttt{CheckCorrectAnswer}(\texttt{gold}, \texttt{valid\_poi\_last\_sentence}), \\
c_{\mathrm{poi}} &= \texttt{CheckCorrectAnswer}(\texttt{gold}, \texttt{poi\_last\_sentence}), \\
c_{\mathrm{last}} &= \texttt{CheckCorrectAnswer}(\texttt{gold}, \texttt{last\_sentence}), \\
a_{\mathrm{valid}} &= \texttt{AltFlag}(\texttt{valid\_poi\_last\_sentence}, \texttt{alts}, \texttt{gold}), \\
a_{\mathrm{poi}} &= \texttt{AltFlag}(\texttt{poi\_last\_sentence}, \texttt{alts}, \texttt{gold}), \\
a_{\mathrm{last}} &= \texttt{AltFlag}(\texttt{last\_sentence}, \texttt{alts}, \texttt{gold}).
\end{align*}

Classification (first applicable wins):
\begin{enumerate}
    \item if \(c_{\mathrm{valid}}\) and not \(a_{\mathrm{valid}}\): \texttt{correct\_valid}
    \item else if \(c_{\mathrm{poi}}\) and not \(a_{\mathrm{poi}}\): \texttt{correct\_poi}
    \item else if \(c_{\mathrm{last}}\) and not \(a_{\mathrm{last}}\): \texttt{correct\_last\_sentence}
    \item else if not \(c_{\mathrm{valid}}\) and \(a_{\mathrm{valid}}\) and \(\texttt{valid\_poi\_last\_sentence} \neq \text{""}\): \texttt{wrong\_logic}
    \item else if not \(c_{\mathrm{poi}}\) and \(a_{\mathrm{poi}}\) and \(\texttt{poi\_last\_sentence} \neq \text{""}\): \texttt{wrong\_logic\_poi}
    \item else if not \(c_{\mathrm{last}}\) and \(a_{\mathrm{last}}\) and \(\texttt{last\_sentence} \neq \text{""}\): \texttt{wrong\_logic\_last\_sentence}
    \item else: \texttt{wrong\_other}
\end{enumerate}

\noindent Note: When a sentence contains both gold and an alternative that spans the gold, this does not trigger a \texttt{wrong\_logic*} bucket unless the corresponding ``not \(c\_\cdot\) and \(a\_\cdot\)'' guard holds; such cases can fall through to \texttt{wrong\_other}, matching the implementation. The main figures/regressions of the paper use accuracy with \(\text{correct} \in \{\texttt{correct\_valid}, \texttt{correct\_poi}, \texttt{correct\_last\_sentence}\}\) and \text{correct} as the complement. 

\subsection{Pseudocode}
Using pseudo code, figure \ref{algo:data-prep} explains the data preparation, figure \ref{algo:scoring} explains the scoring of the results and success/failure types, and figure  \ref{algo:utilities} details the used utility functions.

\begin{algorithm}[h]
\DontPrintSemicolon
\SetKwProg{Fn}{Procedure}{}{}
\SetKwFunction{Eval}{EvaluateOutput}
\SetKwFunction{MapCat}{MapQuestionToCategory}
\SetKwFunction{Quals}{GetQualifiersFor}
\SetKwFunction{ExCtx}{ExceededContext}
\SetKwFunction{TLS}{TakeLastSentence}
\SetKwFunction{CCA}{CheckCorrectAnswer}
\SetKwFunction{AF}{AltFlag}

\tcp{Parse identifiers and gold from puzzle name: N\_d\_needles\_puzzleId\_gold}
\((N, d, \text{needles}, \text{puzzleId}, \text{gold}) \leftarrow \textsc{parse}(\texttt{ans\_path})\);\;

\tcp{Load question and resolve category}
\(\text{question} \leftarrow \textsc{last\_line}(\texttt{puzzle\_txt\_path})\);\;
\(\text{category} \leftarrow \MapCat(\text{question})\);\;

\tcp{Load metadata needed for evaluation}
\((\text{poi}, \text{domains}) \leftarrow \textsc{readJSON}(\texttt{state\_json\_path})\);\;
\(\text{poi} \leftarrow \textsc{lower}(\text{poi}); \; \text{domainValues} \leftarrow \text{domains}[\text{category}]\);\;
\(\text{quals} \leftarrow \Quals(\text{category})\);\;

\tcp{Read and normalize model output}
\(\text{resp} \leftarrow \textsc{read}(\texttt{ans\_path})\);\;
\(\text{lines} \leftarrow \textsc{split\_lines}(\textsc{lower}(\text{resp}))\);\;
\If{\(\textsc{startsWith}(\text{lines}[-1], "(") \land \textsc{endsWith}(\text{lines}[-1], ")")\)}{\(\textsc{drop\_last}(\text{lines})\)}\;
\(\text{lines} \leftarrow \textsc{filter\_nonempty}(\text{lines})\);\;
\If{\(\textsc{len}(\text{lines})=0\)}{\(\text{lines}\leftarrow[""]\)}\;
\caption{Pseudocode of the data preparation.}
\label{algo:data-prep}
\end{algorithm}

\begin{algorithm}[h]
\DontPrintSemicolon
\SetKwProg{Fn}{Procedure}{}{}
\SetKwFunction{Eval}{EvaluateOutput}
\SetKwFunction{MapCat}{MapQuestionToCategory}
\SetKwFunction{Quals}{GetQualifiersFor}
\SetKwFunction{ExCtx}{ExceededContext}
\SetKwFunction{TLS}{TakeLastSentence}
\SetKwFunction{CCA}{CheckCorrectAnswer}
\SetKwFunction{AF}{AltFlag}
\tcp{Prepare data}
\tcp{Extract windows}
\(\text{last\_sentence} \leftarrow \TLS(\text{lines}[-1])\);\;
\(\text{poi\_lines} \leftarrow \{s\in \text{lines} : \textsc{contains}(s, \text{poi})\}\);\;
\(\text{poi\_last} \leftarrow \textsc{lastOrEmpty}(\text{poi\_lines})\);\;
\(\text{valid\_poi\_lines} \leftarrow \{s\in \text{poi\_lines} : \textsc{containsAny}(s, \text{quals})\}\);\;
\(\text{valid\_poi\_last} \leftarrow \textsc{lastOrEmpty}(\text{valid\_poi\_lines})\);\;

\tcp{Budget/emptiness guard}
\If{\(\ExCtx(\text{runtime\_meta}) \lor \text{last\_sentence} = "" \lor \textsc{len}(\text{resp})=0\)}{
    \Return \texttt{wrong\_max\_context}
}

\tcp{Gold matches}
\(c_{\mathrm{valid}} \leftarrow \CCA(\text{gold}, \text{valid\_poi\_last})\);\;
\(c_{\mathrm{poi}} \leftarrow \CCA(\text{gold}, \text{poi\_last})\);\;
\(c_{\mathrm{last}} \leftarrow \CCA(\text{gold}, \text{last\_sentence})\);\;

\tcp{Alternative flags with span rule}
\(\text{alts} \leftarrow \{v\in \text{domainValues} : v \neq \text{gold}\}\);\;
\(a_{\mathrm{valid}} \leftarrow \AF(\text{valid\_poi\_last}, \text{alts}, \text{gold})\);\;
\(a_{\mathrm{poi}} \leftarrow \AF(\text{poi\_last}, \text{alts}, \text{gold})\);\;
\(a_{\mathrm{last}} \leftarrow \AF(\text{last\_sentence}, \text{alts}, \text{gold})\);\;

\tcp{Buckets (first match wins)}
\If{\(c_{\mathrm{valid}} \land \lnot a_{\mathrm{valid}}\)}{\Return \texttt{correct\_valid}}
\ElseIf{\(c_{\mathrm{poi}} \land \lnot a_{\mathrm{poi}}\)}{\Return \texttt{correct\_poi}}
\ElseIf{\(c_{\mathrm{last}} \land \lnot a_{\mathrm{last}}\)}{\Return \texttt{correct\_last\_sentence}}
\ElseIf{\(\lnot c_{\mathrm{valid}} \land a_{\mathrm{valid}} \land \text{valid\_poi\_last}\neq ""\)}{\Return \texttt{wrong\_logic}}
\ElseIf{\(\lnot c_{\mathrm{poi}} \land a_{\mathrm{poi}} \land \text{poi\_last}\neq ""\)}{\Return \texttt{wrong\_logic\_poi}}
\ElseIf{\(\lnot c_{\mathrm{last}} \land a_{\mathrm{last}} \land \text{last\_sentence}\neq ""\)}{\Return \texttt{wrong\_logic\_last\_sentence}}
\Else{\Return \texttt{wrong\_other}}

\caption{Pseudocode of the CogniLoad scoring pipeline.}

\label{algo:scoring}
\end{algorithm}

\begin{algorithm}[h]
\DontPrintSemicolon
\SetKwProg{Fn}{Procedure}{}{}
\SetKwFunction{Eval}{EvaluateOutput}
\SetKwFunction{MapCat}{MapQuestionToCategory}
\SetKwFunction{Quals}{GetQualifiersFor}
\SetKwFunction{ExCtx}{ExceededContext}
\SetKwFunction{TLS}{TakeLastSentence}
\SetKwFunction{CCA}{CheckCorrectAnswer}
\SetKwFunction{AF}{AltFlag}

\textbf{Utilities.}

\textbf{ExceededContext}: return true if \(\texttt{prompt\_tokens} + \texttt{response\_tokens} + 20 \ge 32~768\).

\textbf{TakeLastSentence}(\(s\)): let \(p=\textsc{split}(s,".")\). If \(\textsc{len}(p)\ge 2\), return \(p[-2]\); else \(s\).

\textbf{CheckCorrectAnswer}(\(gold, s\)):
\begin{itemize}
    \item \textbf{CheckAnswer}(\(t, s\)) is true if any holds: \(\textsc{startsWith}(s,t)\), \(s=t\), or \(s\) contains one of: ``\,space\(+t\)\,'', ``[\,\(t\)'', ``" \(t\)'', ``*\(t\)'', ``\_\(t\)'', ``\{ \(t\)'', ``(\(t\)''.
    \item If \(gold=\) \texttt{reaggea}: return \(\textbf{CheckAnswer}(\text{reaggea}) \lor \textbf{CheckAnswer}(\text{reggae})\).
    \item If \texttt{sci-fi}: return \(\textbf{CheckAnswer}(\text{sci-fi}) \lor \textbf{CheckAnswer}(\text{science fiction}) \lor \textbf{CheckAnswer}(\text{science-fiction})\).
    \item If \texttt{camp}: \(\textbf{CheckAnswer}(\text{camp}) \lor \textbf{CheckAnswer}(\text{campground})\).
    \item If \texttt{potatoes}: \(\textbf{CheckAnswer}(\text{potatoes}) \lor \textbf{CheckAnswer}(\text{potato})\).
    \item If \texttt{market}: \(\textbf{CheckAnswer}(\text{market}) \lor \textbf{CheckAnswer}(\text{marketplace})\).
    \item If \texttt{livingroom}: \(\textbf{CheckAnswer}(\text{livingroom}) \lor \textbf{CheckAnswer}(\text{living room})\).
    \item Else: \(\textbf{CheckAnswer}(\text{gold})\).
\end{itemize}

\textbf{AltFlag}(\(s, \text{alts}, \text{gold}\)): 
\begin{enumerate}
    \item \(s \leftarrow \textsc{lower}(s)\), \(g \leftarrow \textsc{lower}(\text{gold})\), \(A \leftarrow \{\textsc{lower}(a) : a\in \text{alts}, a\neq g\}\).
    \item \(\text{alt\_present} \leftarrow \bigvee_{a\in A} \textbf{CheckCorrectAnswer}(a, s)\).
    \item \(g\_idx \leftarrow s.\textsc{rfind}(g)\).
    \item \textbf{if} \(g\_idx\neq -1\) and \(\text{alt\_present}\): let \(g\_end=g\_idx+\textsc{len}(g)\); set \(\text{spans}=\)false. For each \(a\in A\): \(a\_idx \leftarrow s.\textsc{rfind}(a)\); if \(a\_idx\neq -1\) and \(a\_idx \le g\_idx\) and \(a\_idx+\textsc{len}(a) \ge g\_end\), set \(\text{spans}=\)true. If \(\lnot \text{spans}\), set \(\text{alt\_present}=\)false.
    \item Return \(\text{alt\_present}\).
\end{enumerate}

\caption{Pseudocode of the CogniLoad scoring pipeline utility functions.}

\label{algo:utilities}
\end{algorithm}

\section{Error distribution}
\label{app:err_dist}
The following tables illustrate the error-type distribution per difficulty \ref{tab:error_by_model_d}, and length \ref{tab:error_by_model_N}. The purpose of this distinction is to illustrate three primary error types (i.e., long context error, logic error, other error) and to distinguish among the logic errors how well the model followed the instruction in its response formatting. The other error captures cases where the model responds but doesn't mention any attribute, usually this is cases where the model erroneously assumes the problem can't be solved with the provided information.  For conciseness the error types in the table headers are abbreviated as following:

\begin{itemize}
    \item Wrong max-context  \(\to\) ctx
    \item Wrong logic (last valid POI sentence) \(\to\) valid-logic
    \item Wrong logic (last POI sentence) \(\to\) poi-logic
    \item Wrong logic (last sentence) \(\to\) last-logic
    \item Wrong other \(\to\) other
\end{itemize}

If the total number of evaluated samples does not precisely match the expected number of samples (e.g `o3-2025-04-16` with difficulty 5 resulted in 279 vs 280 expected examples), this usually indicates the puzzle was not processed due to a an API error (e.g. a security refusal). These few missing cases were attempted to be evaluated a second time but refused again.

\begin{longtable}{lrrrrrrrr}
\caption{Error-type distribution by difficulty $d$ per model (aggregated over $N$ and $\rho$).}\label{tab:error_by_model_d}\\
\toprule
Model & Dim & Samples & Acc [\%] & ctx & valid-logic & poi-logic & last-logic & other \\
\midrule
\endfirsthead
\caption[]{Error-type distribution by difficulty $d$ per model (aggregated over $N$ and $rho$). (continued)}\\
\toprule
Model & Dim & Samples & Acc [\%] & ctx & valid-logic & poi-logic & last-logic & other \\
\midrule
\endhead
\bottomrule
\endfoot
DeepSeek-R1-0528 & 1 & 280 & 88 & 3 & 31 & 0 & 0 & 0 \\
DeepSeek-R1-0528 & 3 & 280 & 78 & 11 & 52 & 0 & 0 & 0 \\
DeepSeek-R1-0528 & 5 & 280 & 67 & 13 & 80 & 0 & 0 & 0 \\
DeepSeek-R1-0528 & 7 & 280 & 55 & 16 & 109 & 0 & 0 & 0 \\
DeepSeek-R1-0528 & 10 & 280 & 48 & 19 & 128 & 0 & 0 & 0 \\
DeepSeek-R1-Distill-Llama-70B & 1 & 2800 & 82 & 0 & 507 & 0 & 2 & 9 \\
DeepSeek-R1-Distill-Llama-70B & 3 & 2800 & 68 & 11 & 877 & 0 & 3 & 3 \\
DeepSeek-R1-Distill-Llama-70B & 5 & 2800 & 54 & 12 & 1260 & 0 & 8 & 2 \\
DeepSeek-R1-Distill-Llama-70B & 7 & 2800 & 48 & 28 & 1424 & 0 & 4 & 4 \\
DeepSeek-R1-Distill-Llama-70B & 10 & 2800 & 36 & 33 & 1747 & 0 & 4 & 5 \\
DeepSeek-R1-Distill-Qwen-1.5B & 1 & 2800 & 31 & 561 & 1140 & 4 & 81 & 154 \\
DeepSeek-R1-Distill-Qwen-1.5B & 3 & 2800 & 19 & 543 & 1305 & 3 & 82 & 347 \\
DeepSeek-R1-Distill-Qwen-1.5B & 5 & 2800 & 12 & 613 & 1381 & 4 & 63 & 413 \\
DeepSeek-R1-Distill-Qwen-1.5B & 7 & 2800 & 10 & 695 & 1323 & 7 & 60 & 443 \\
DeepSeek-R1-Distill-Qwen-1.5B & 10 & 2800 & 8 & 717 & 1371 & 7 & 62 & 407 \\
DeepSeek-R1-Distill-Qwen-32B & 1 & 2800 & 63 & 1 & 1023 & 0 & 3 & 5 \\
DeepSeek-R1-Distill-Qwen-32B & 3 & 2800 & 66 & 46 & 908 & 0 & 2 & 1 \\
DeepSeek-R1-Distill-Qwen-32B & 5 & 2800 & 52 & 78 & 1275 & 0 & 1 & 2 \\
DeepSeek-R1-Distill-Qwen-32B & 7 & 2800 & 42 & 94 & 1532 & 0 & 0 & 3 \\
DeepSeek-R1-Distill-Qwen-32B & 10 & 2800 & 31 & 106 & 1813 & 0 & 0 & 1 \\
DeepSeek-R1-Distill-Qwen-7B & 1 & 2800 & 46 & 47 & 1454 & 5 & 4 & 15 \\
DeepSeek-R1-Distill-Qwen-7B & 3 & 2800 & 32 & 124 & 1561 & 5 & 70 & 134 \\
DeepSeek-R1-Distill-Qwen-7B & 5 & 2800 & 20 & 246 & 1706 & 3 & 65 & 206 \\
DeepSeek-R1-Distill-Qwen-7B & 7 & 2800 & 14 & 317 & 1802 & 8 & 49 & 223 \\
DeepSeek-R1-Distill-Qwen-7B & 10 & 2800 & 12 & 352 & 1770 & 6 & 60 & 290 \\
EXAONE-Deep-32B & 1 & 2800 & 62 & 0 & 995 & 5 & 2 & 48 \\
EXAONE-Deep-32B & 3 & 2800 & 42 & 2 & 1595 & 2 & 27 & 10 \\
EXAONE-Deep-32B & 5 & 2800 & 29 & 10 & 1938 & 2 & 45 & 5 \\
EXAONE-Deep-32B & 7 & 2800 & 24 & 14 & 2060 & 1 & 42 & 9 \\
EXAONE-Deep-32B & 10 & 2800 & 17 & 16 & 2257 & 1 & 40 & 13 \\
gemini-2.5-flash & 1 & 280 & 89 & 6 & 24 & 0 & 0 & 0 \\
gemini-2.5-flash & 3 & 280 & 71 & 64 & 17 & 0 & 0 & 0 \\
gemini-2.5-flash & 5 & 280 & 63 & 93 & 11 & 0 & 0 & 0 \\
gemini-2.5-flash & 7 & 280 & 56 & 116 & 8 & 0 & 0 & 0 \\
gemini-2.5-flash & 10 & 280 & 52 & 110 & 25 & 0 & 0 & 0 \\
gemini-2.5-flash-lite & 1 & 280 & 68 & 0 & 91 & 0 & 0 & 0 \\
gemini-2.5-flash-lite & 3 & 280 & 48 & 5 & 138 & 0 & 0 & 2 \\
gemini-2.5-flash-lite & 5 & 280 & 36 & 3 & 176 & 0 & 0 & 1 \\
gemini-2.5-flash-lite & 7 & 280 & 22 & 8 & 209 & 0 & 0 & 2 \\
gemini-2.5-flash-lite & 10 & 280 & 16 & 1 & 227 & 2 & 0 & 4 \\
gemini-2.5-pro & 1 & 280 & 95 & 10 & 5 & 0 & 0 & 0 \\
gemini-2.5-pro & 3 & 280 & 74 & 69 & 3 & 0 & 0 & 0 \\
gemini-2.5-pro & 5 & 280 & 72 & 71 & 8 & 0 & 0 & 0 \\
gemini-2.5-pro & 7 & 280 & 69 & 79 & 7 & 0 & 0 & 0 \\
gemini-2.5-pro & 10 & 280 & 65 & 88 & 9 & 0 & 0 & 0 \\
gpt-5-2025-08-07 & 1 & 280 & 100 & 0 & 1 & 0 & 0 & 0 \\
gpt-5-2025-08-07 & 3 & 280 & 98 & 0 & 7 & 0 & 0 & 0 \\
gpt-5-2025-08-07 & 5 & 280 & 94 & 0 & 17 & 0 & 0 & 0 \\
gpt-5-2025-08-07 & 7 & 280 & 89 & 3 & 26 & 0 & 0 & 1 \\
gpt-5-2025-08-07 & 10 & 280 & 82 & 29 & 21 & 0 & 0 & 0 \\
gpt-5-mini-2025-08-07 & 1 & 280 & 92 & 0 & 14 & 0 & 0 & 7 \\
gpt-5-mini-2025-08-07 & 3 & 280 & 83 & 0 & 46 & 0 & 0 & 2 \\
gpt-5-mini-2025-08-07 & 5 & 280 & 78 & 3 & 52 & 0 & 1 & 5 \\
gpt-5-mini-2025-08-07 & 7 & 280 & 74 & 12 & 58 & 0 & 0 & 2 \\
gpt-5-mini-2025-08-07 & 10 & 280 & 68 & 27 & 63 & 0 & 0 & 1 \\
gpt-5-nano-2025-08-07 & 1 & 280 & 76 & 0 & 65 & 1 & 0 & 0 \\
gpt-5-nano-2025-08-07 & 3 & 280 & 51 & 0 & 133 & 0 & 3 & 0 \\
gpt-5-nano-2025-08-07 & 5 & 280 & 48 & 0 & 145 & 0 & 2 & 0 \\
gpt-5-nano-2025-08-07 & 7 & 280 & 34 & 0 & 183 & 0 & 1 & 0 \\
gpt-5-nano-2025-08-07 & 10 & 280 & 21 & 0 & 219 & 0 & 1 & 0 \\
o3-2025-04-16 & 1 & 279 & 96 & 0 & 11 & 0 & 0 & 0 \\
o3-2025-04-16 & 3 & 280 & 93 & 0 & 20 & 0 & 0 & 0 \\
o3-2025-04-16 & 5 & 279 & 90 & 1 & 28 & 0 & 0 & 0 \\
o3-2025-04-16 & 7 & 280 & 89 & 2 & 30 & 0 & 0 & 0 \\
o3-2025-04-16 & 10 & 280 & 80 & 21 & 34 & 0 & 0 & 0 \\
o4-mini-2025-04-16 & 1 & 279 & 88 & 0 & 31 & 0 & 0 & 2 \\
o4-mini-2025-04-16 & 3 & 280 & 81 & 0 & 53 & 0 & 0 & 0 \\
o4-mini-2025-04-16 & 5 & 280 & 79 & 3 & 57 & 0 & 0 & 0 \\
o4-mini-2025-04-16 & 7 & 279 & 70 & 5 & 78 & 0 & 0 & 1 \\
o4-mini-2025-04-16 & 10 & 280 & 65 & 13 & 86 & 0 & 0 & 0 \\
Phi-4-mini-reasoning & 1 & 2800 & 49 & 4 & 1222 & 16 & 48 & 126 \\
Phi-4-mini-reasoning & 3 & 2800 & 31 & 1 & 1613 & 6 & 126 & 178 \\
Phi-4-mini-reasoning & 5 & 2800 & 20 & 12 & 1733 & 8 & 286 & 189 \\
Phi-4-mini-reasoning & 7 & 2800 & 16 & 7 & 1785 & 6 & 354 & 203 \\
Phi-4-mini-reasoning & 10 & 2800 & 11 & 18 & 1856 & 5 & 415 & 187 \\
Phi-4-reasoning & 1 & 2800 & 83 & 14 & 417 & 0 & 0 & 42 \\
Phi-4-reasoning & 3 & 2800 & 63 & 79 & 921 & 0 & 0 & 33 \\
Phi-4-reasoning & 5 & 2800 & 47 & 246 & 1219 & 1 & 0 & 30 \\
Phi-4-reasoning & 7 & 2800 & 36 & 522 & 1240 & 0 & 0 & 29 \\
Phi-4-reasoning & 10 & 2800 & 26 & 857 & 1200 & 1 & 0 & 25 \\
Phi-4-reasoning-plus & 1 & 2800 & 83 & 100 & 347 & 1 & 0 & 37 \\
Phi-4-reasoning-plus & 3 & 2800 & 58 & 182 & 938 & 0 & 0 & 43 \\
Phi-4-reasoning-plus & 5 & 2800 & 40 & 471 & 1153 & 1 & 0 & 46 \\
Phi-4-reasoning-plus & 7 & 2800 & 30 & 936 & 1001 & 2 & 0 & 27 \\
Phi-4-reasoning-plus & 10 & 2800 & 21 & 1198 & 994 & 0 & 1 & 32 \\
Qwen3-1.7B & 1 & 2800 & 33 & 1 & 1840 & 1 & 11 & 17 \\
Qwen3-1.7B & 3 & 2800 & 28 & 0 & 2010 & 0 & 2 & 11 \\
Qwen3-1.7B & 5 & 2800 & 19 & 0 & 2268 & 0 & 1 & 8 \\
Qwen3-1.7B & 7 & 2800 & 14 & 0 & 2396 & 0 & 2 & 10 \\
Qwen3-1.7B & 10 & 2798 & 11 & 0 & 2483 & 0 & 1 & 5 \\
Qwen3-30B-A3B & 1 & 2800 & 75 & 18 & 652 & 3 & 0 & 38 \\
Qwen3-30B-A3B & 3 & 2800 & 54 & 48 & 1237 & 1 & 0 & 6 \\
Qwen3-30B-A3B & 5 & 2800 & 41 & 40 & 1587 & 2 & 0 & 10 \\
Qwen3-30B-A3B & 7 & 2800 & 31 & 25 & 1894 & 2 & 0 & 6 \\
Qwen3-30B-A3B & 10 & 2800 & 23 & 19 & 2133 & 3 & 1 & 8 \\
Qwen3-32B & 1 & 2800 & 76 & 0 & 626 & 0 & 1 & 49 \\
Qwen3-32B & 3 & 2800 & 61 & 1 & 1075 & 1 & 1 & 10 \\
Qwen3-32B & 5 & 2800 & 48 & 0 & 1434 & 1 & 3 & 6 \\
Qwen3-32B & 7 & 2800 & 38 & 0 & 1721 & 0 & 6 & 9 \\
Qwen3-32B & 10 & 2800 & 28 & 1 & 1994 & 0 & 12 & 10 \\
Qwen3-8B & 1 & 2800 & 68 & 77 & 739 & 2 & 0 & 76 \\
Qwen3-8B & 3 & 2800 & 50 & 136 & 1190 & 3 & 18 & 49 \\
Qwen3-8B & 5 & 2800 & 36 & 106 & 1619 & 3 & 19 & 37 \\
Qwen3-8B & 7 & 2800 & 30 & 101 & 1791 & 0 & 27 & 39 \\
Qwen3-8B & 10 & 2800 & 22 & 78 & 2046 & 0 & 39 & 25 \\
QwQ-32B & 1 & 2800 & 62 & 2 & 1044 & 1 & 0 & 5 \\
QwQ-32B & 3 & 2800 & 69 & 12 & 850 & 0 & 3 & 1 \\
QwQ-32B & 5 & 2800 & 52 & 52 & 1279 & 0 & 3 & 6 \\
QwQ-32B & 7 & 2800 & 39 & 112 & 1596 & 0 & 1 & 2 \\
QwQ-32B & 10 & 2800 & 30 & 273 & 1674 & 0 & 5 & 9 \\
\end{longtable}

\begin{longtable}{lrrrrrrrr}
\caption{Error-type distribution by puzzle length $N$ per model (aggregated over $d$ and $\rho$).}\label{tab:error_by_model_N}\\
\toprule
Model & Dim & Samples & Acc [\%] & ctx & valid-logic & poi-logic & last-logic & other \\
\midrule
\endfirsthead
\caption[]{Error-type distribution by puzzle length $N$ per model (aggregated over $d$ and $\rho$). (continued)}\\
\toprule
Model & Dim & Samples & Acc [\%] & ctx & valid-logic & poi-logic & last-logic & other \\
\midrule
\endhead
\bottomrule
\endfoot
DeepSeek-R1-0528 & 20 & 350 & 97 & 7 & 2 & 0 & 0 & 0 \\
DeepSeek-R1-0528 & 50 & 350 & 87 & 14 & 30 & 0 & 0 & 0 \\
DeepSeek-R1-0528 & 100 & 350 & 56 & 25 & 128 & 0 & 0 & 0 \\
DeepSeek-R1-0528 & 250 & 350 & 27 & 16 & 240 & 0 & 0 & 0 \\
DeepSeek-R1-Distill-Llama-70B & 20 & 3500 & 89 & 0 & 372 & 0 & 1 & 4 \\
DeepSeek-R1-Distill-Llama-70B & 50 & 3500 & 66 & 2 & 1167 & 0 & 4 & 5 \\
DeepSeek-R1-Distill-Llama-70B & 100 & 3500 & 48 & 2 & 1811 & 0 & 6 & 3 \\
DeepSeek-R1-Distill-Llama-70B & 250 & 3500 & 27 & 80 & 2465 & 0 & 10 & 11 \\
DeepSeek-R1-Distill-Qwen-1.5B & 20 & 3500 & 19 & 749 & 1676 & 8 & 58 & 349 \\
DeepSeek-R1-Distill-Qwen-1.5B & 50 & 3500 & 16 & 752 & 1749 & 5 & 85 & 360 \\
DeepSeek-R1-Distill-Qwen-1.5B & 100 & 3500 & 15 & 698 & 1749 & 4 & 90 & 450 \\
DeepSeek-R1-Distill-Qwen-1.5B & 250 & 3500 & 14 & 930 & 1346 & 8 & 115 & 605 \\
DeepSeek-R1-Distill-Qwen-32B & 20 & 3500 & 78 & 1 & 760 & 0 & 0 & 2 \\
DeepSeek-R1-Distill-Qwen-32B & 50 & 3500 & 55 & 1 & 1583 & 0 & 2 & 3 \\
DeepSeek-R1-Distill-Qwen-32B & 100 & 3500 & 41 & 0 & 2061 & 0 & 1 & 3 \\
DeepSeek-R1-Distill-Qwen-32B & 250 & 3500 & 29 & 323 & 2147 & 0 & 3 & 4 \\
DeepSeek-R1-Distill-Qwen-7B & 20 & 3500 & 37 & 117 & 2012 & 4 & 14 & 63 \\
DeepSeek-R1-Distill-Qwen-7B & 50 & 3500 & 24 & 207 & 2298 & 1 & 43 & 123 \\
DeepSeek-R1-Distill-Qwen-7B & 100 & 3500 & 20 & 311 & 2177 & 11 & 75 & 221 \\
DeepSeek-R1-Distill-Qwen-7B & 250 & 3500 & 19 & 451 & 1806 & 11 & 116 & 461 \\
EXAONE-Deep-32B & 20 & 3500 & 57 & 7 & 1451 & 4 & 35 & 20 \\
EXAONE-Deep-32B & 50 & 3500 & 33 & 17 & 2255 & 2 & 56 & 19 \\
EXAONE-Deep-32B & 100 & 3500 & 25 & 7 & 2556 & 4 & 34 & 21 \\
EXAONE-Deep-32B & 250 & 3500 & 24 & 11 & 2583 & 1 & 31 & 25 \\
gemini-2.5-flash & 20 & 350 & 99 & 0 & 4 & 0 & 0 & 0 \\
gemini-2.5-flash & 50 & 350 & 91 & 10 & 22 & 0 & 0 & 0 \\
gemini-2.5-flash & 100 & 350 & 61 & 99 & 39 & 0 & 0 & 0 \\
gemini-2.5-flash & 250 & 350 & 14 & 280 & 20 & 0 & 0 & 0 \\
gemini-2.5-flash-lite & 20 & 350 & 55 & 0 & 154 & 0 & 0 & 2 \\
gemini-2.5-flash-lite & 50 & 350 & 35 & 0 & 223 & 1 & 0 & 3 \\
gemini-2.5-flash-lite & 100 & 350 & 32 & 1 & 235 & 0 & 0 & 2 \\
gemini-2.5-flash-lite & 250 & 350 & 29 & 16 & 229 & 1 & 0 & 2 \\
gemini-2.5-pro & 20 & 350 & 99 & 0 & 2 & 0 & 0 & 0 \\
gemini-2.5-pro & 50 & 350 & 98 & 1 & 5 & 0 & 0 & 0 \\
gemini-2.5-pro & 100 & 350 & 82 & 48 & 15 & 0 & 0 & 0 \\
gemini-2.5-pro & 250 & 350 & 21 & 268 & 10 & 0 & 0 & 0 \\
gpt-5-2025-08-07 & 20 & 350 & 100 & 0 & 1 & 0 & 0 & 0 \\
gpt-5-2025-08-07 & 50 & 350 & 98 & 0 & 7 & 0 & 0 & 0 \\
gpt-5-2025-08-07 & 100 & 350 & 96 & 0 & 12 & 0 & 0 & 1 \\
gpt-5-2025-08-07 & 250 & 350 & 76 & 32 & 52 & 0 & 0 & 0 \\
gpt-5-mini-2025-08-07 & 20 & 350 & 96 & 0 & 12 & 0 & 0 & 3 \\
gpt-5-mini-2025-08-07 & 50 & 350 & 93 & 0 & 23 & 0 & 0 & 3 \\
gpt-5-mini-2025-08-07 & 100 & 350 & 84 & 0 & 54 & 0 & 0 & 3 \\
gpt-5-mini-2025-08-07 & 250 & 350 & 44 & 42 & 144 & 0 & 1 & 8 \\
gpt-5-nano-2025-08-07 & 20 & 350 & 75 & 0 & 86 & 0 & 0 & 0 \\
gpt-5-nano-2025-08-07 & 50 & 350 & 48 & 0 & 181 & 1 & 1 & 0 \\
gpt-5-nano-2025-08-07 & 100 & 350 & 37 & 0 & 217 & 0 & 3 & 0 \\
gpt-5-nano-2025-08-07 & 250 & 350 & 25 & 0 & 261 & 0 & 3 & 0 \\
o3-2025-04-16 & 20 & 350 & 99 & 0 & 5 & 0 & 0 & 0 \\
o3-2025-04-16 & 50 & 350 & 98 & 0 & 8 & 0 & 0 & 0 \\
o3-2025-04-16 & 100 & 350 & 94 & 0 & 22 & 0 & 0 & 0 \\
o3-2025-04-16 & 250 & 348 & 68 & 24 & 88 & 0 & 0 & 0 \\
o4-mini-2025-04-16 & 20 & 350 & 99 & 0 & 3 & 0 & 0 & 1 \\
o4-mini-2025-04-16 & 50 & 350 & 93 & 0 & 24 & 0 & 0 & 1 \\
o4-mini-2025-04-16 & 100 & 350 & 76 & 0 & 85 & 0 & 0 & 0 \\
o4-mini-2025-04-16 & 250 & 348 & 38 & 21 & 193 & 0 & 0 & 1 \\
Phi-4-mini-reasoning & 20 & 3500 & 37 & 8 & 1765 & 7 & 213 & 215 \\
Phi-4-mini-reasoning & 50 & 3500 & 24 & 4 & 2163 & 11 & 289 & 208 \\
Phi-4-mini-reasoning & 100 & 3500 & 21 & 13 & 2198 & 6 & 327 & 222 \\
Phi-4-mini-reasoning & 250 & 3500 & 21 & 17 & 2083 & 17 & 400 & 238 \\
Phi-4-reasoning & 20 & 3500 & 84 & 11 & 496 & 0 & 0 & 38 \\
Phi-4-reasoning & 50 & 3500 & 56 & 106 & 1393 & 0 & 0 & 43 \\
Phi-4-reasoning & 100 & 3500 & 40 & 332 & 1736 & 1 & 0 & 43 \\
Phi-4-reasoning & 250 & 3500 & 24 & 1269 & 1372 & 1 & 0 & 35 \\
Phi-4-reasoning-plus & 20 & 3500 & 81 & 12 & 586 & 1 & 0 & 55 \\
Phi-4-reasoning-plus & 50 & 3500 & 50 & 382 & 1326 & 3 & 0 & 54 \\
Phi-4-reasoning-plus & 100 & 3500 & 35 & 668 & 1563 & 0 & 1 & 45 \\
Phi-4-reasoning-plus & 250 & 3500 & 20 & 1825 & 958 & 0 & 0 & 31 \\
Qwen3-1.7B & 20 & 3500 & 27 & 0 & 2546 & 0 & 5 & 12 \\
Qwen3-1.7B & 50 & 3500 & 20 & 1 & 2796 & 1 & 3 & 13 \\
Qwen3-1.7B & 100 & 3500 & 19 & 0 & 2818 & 0 & 6 & 8 \\
Qwen3-1.7B & 250 & 3498 & 18 & 0 & 2837 & 0 & 3 & 18 \\
Qwen3-30B-A3B & 20 & 3500 & 72 & 25 & 942 & 1 & 1 & 12 \\
Qwen3-30B-A3B & 50 & 3500 & 48 & 40 & 1764 & 2 & 0 & 19 \\
Qwen3-30B-A3B & 100 & 3500 & 34 & 44 & 2228 & 1 & 0 & 21 \\
Qwen3-30B-A3B & 250 & 3500 & 25 & 41 & 2569 & 7 & 0 & 16 \\
Qwen3-32B & 20 & 3500 & 82 & 0 & 607 & 0 & 1 & 8 \\
Qwen3-32B & 50 & 3500 & 55 & 1 & 1577 & 0 & 3 & 10 \\
Qwen3-32B & 100 & 3500 & 38 & 0 & 2125 & 0 & 4 & 35 \\
Qwen3-32B & 250 & 3500 & 26 & 1 & 2541 & 2 & 15 & 31 \\
Qwen3-8B & 20 & 3500 & 71 & 75 & 912 & 0 & 18 & 24 \\
Qwen3-8B & 50 & 3500 & 40 & 108 & 1893 & 1 & 31 & 53 \\
Qwen3-8B & 100 & 3500 & 31 & 117 & 2198 & 3 & 27 & 64 \\
Qwen3-8B & 250 & 3500 & 23 & 198 & 2382 & 4 & 27 & 85 \\
QwQ-32B & 20 & 3500 & 76 & 8 & 810 & 0 & 4 & 7 \\
QwQ-32B & 50 & 3500 & 55 & 27 & 1550 & 0 & 1 & 6 \\
QwQ-32B & 100 & 3500 & 42 & 36 & 1991 & 0 & 5 & 5 \\
QwQ-32B & 250 & 3500 & 29 & 380 & 2092 & 1 & 2 & 5 \\
\end{longtable}

\section{AIC-Comparison}
\label{app:aic}
In this section we compare the Akaike Information Criterion (AIC) of the GLM model containing a pure linear $\rho$ term with a GLM model with a squared term added for $\rho$.  The equations for the models are as following with $Y=1$ indicating a correctly solved puzzle:

\paragraph{Linear model } 
\[
\Pr\!\bigl(Y{=}1\bigr) = \sigma\!\bigl(
  \beta_{0}
+ \beta_{d}\,d
+ \beta_{N}\,\log_{10}N
+ \beta_{\rho}\,\rho
\bigr),
\]

\paragraph{Quadratic model}
\[
\Pr\!\bigl(Y{=}1\bigr) = \sigma\!\bigl(
  \beta_{0}
+ \beta_{d}\,d
+ \beta_{N}\,\log_{10}N
+ \beta_{\rho}\,\rho
+ \beta_{\rho^2}\,\rho^2
\bigr),
\]

To assess whether the quadratic specification provides a statistically significant improvement over the linear GLM, we compare their maximised log-likelihoods \(\ell_{\text{lin}}\) and \(\ell_{\text{quad}}\).  The likelihood-ratio statistic \(D = 2(\ell_{\text{quad}}-\ell_{\text{lin}})\) follows, under the null hypothesis that the extra term is unnecessary, a chi-squared distribution with one degree of freedom because the quadratic model introduces exactly one additional parameter.  

The \(p\)-value reported in Table \ref{tab:quad-vs-linear} is the upper-tail probability \(p = \Pr(\chi^{2}_{1} \ge D)\). p-values below \(0.05\) indicate that the quadratic term yields a statistically significant gain in fit and therefore justifies its inclusion.

We find that for all except for two models (i.e., DeepSeek-R1-Distill-Qwen-1.5B and DeepSeek-R1-Distill-Qwen-7B), the quadratic term for $\rho$ results in a significantly improved AIC value. Therefore we include the quadratic $\rho$ term in the GLM specification within the paper.

\begin{table}[ht]
\centering
\begin{tabular}{lccc}
\toprule
model & $p_{\mathrm{LR}}$ & AIC$_{\text{linear}}$ & AIC$_{\text{quad}}$ \\
\midrule
DeepSeek-R1-0528 & \textbf{0.000} & 1032.432 & \textbf{1019.177} \\
DeepSeek-R1-Distill-Llama-70B & \textbf{0.000} & 13888.760 & \textbf{13703.165} \\
DeepSeek-R1-Distill-Qwen-1.5B & 0.657 & 11582.864 & 11584.667 \\
DeepSeek-R1-Distill-Qwen-32B & \textbf{0.000} & 16487.488 & \textbf{16409.647} \\
DeepSeek-R1-Distill-Qwen-7B & \textbf{0.041} & 14227.155 & \textbf{14224.975} \\
EXAONE-Deep-32B & \textbf{0.000} & 15469.929 & \textbf{15388.288} \\
Phi-4-mini-reasoning & \textbf{0.000} & 14462.835 & \textbf{14438.087} \\
Phi-4-reasoning & \textbf{0.000} & 13344.906 & \textbf{13161.306} \\
Phi-4-reasoning-plus & \textbf{0.000} & 12603.062 & \textbf{12477.924} \\
QwQ-32B & \textbf{0.000} & 16357.613 & \textbf{16218.181} \\
Qwen3-1.7B & \textbf{0.000} & 13700.396 & \textbf{13684.334} \\
Qwen3-30B-A3B & \textbf{0.000} & 15363.614 & \textbf{15252.219} \\
Qwen3-32B & \textbf{0.000} & 14818.588 & \textbf{14720.097} \\
Qwen3-8B & \textbf{0.000} & 15522.392 & \textbf{15425.101} \\
gemini-2.5-flash & 0.077 & 807.553 & 806.420 \\
gemini-2.5-flash-lite & \textbf{0.000} & 1574.306 & \textbf{1555.886} \\
gemini-2.5-pro & 0.201 & 642.222 & 642.583 \\
gpt-5-2025-08-07 & \textbf{0.001} & 484.186 & \textbf{474.503} \\
gpt-5-mini-2025-08-07 & \textbf{0.000} & 1044.689 & \textbf{1023.604} \\
gpt-5-nano-2025-08-07 & \textbf{0.000} & 1520.921 & \textbf{1500.138} \\
o3-2025-04-16 & 0.121 & 692.757 & 692.356 \\
o4-mini-2025-04-16 & \textbf{0.000} & 1002.927 & \textbf{970.994} \\
\bottomrule
\end{tabular}
\caption{Model comparison: linear vs.\ quadratic fit. A bold number indicates a significant improvement of AIC$_{\text{quad}}$ over AIC$_{\text{linear}}$ with p $<$ 0.05.}
\label{tab:quad-vs-linear}
\end{table}

% Recommended preamble packages:
% \usepackage{amsmath, amssymb}
% \usepackage[ruled,vlined]{algorithm2e}
% \usepackage{booktabs}

\section{Complete Attribute Ontology}
\label{app:ontology}
The following section lists the full attribute ontology of all values available for the categories.

\paragraph{people}
Peter, Paul, Mary, John, Mark, Jeff, Craig, Daniel, Anna, Arnoldo, Ali, Benjamin, Joe, Donald, Mitch, Chuck, Jack, Lucas, Jeniffer, Adam, Greg, Allan, David, Ellen, Fred, Hank, Hubert, Ian, Ingrid, Rebecca, Ken, Lewis, Michael, Nathaniel, Oliver, Russ, Steve, Sandy, Ted, Tanya, Veronica, Vincent, Wesley, Brad, Sam, Igor, Sue, Jan, Jeffrey, Jacques, Debby, Olivia, Benedict, Chris, Charles, Harry, Eli, Mahmoud, Chen, William, Linda, Elizabeth, Robert, Jennifer, Emily, Joseph, Thomas, Patricia, Anthony, Jessica, Brian, Lisa, Kevin, Karen, Laura, Eric, Stephanie, Michelle, George, Andrew, Joshua, Amber, Timothy, Victoria, Richard, Cynthia, Brandon, Megan, Matthew, Nancy, Jacqueline, Gary, Dorothy, Edward, Kimberly, Scott, Sara, Justin, Brittany, Ronald, Deborah, Janet, Christopher, Alexander, Samantha, Oscar, Cindy, Frank, Carl, Paula, Irene, Theresa, Dennis, Ralph, Gerald, Martin, Terry, Bryan, Lance, Corey, Casey, Brent, Derek, Travis, Austin, Victor, Jesse, Zachary, Kyle, Aaron, Betty, Connie, Holly, Donna, Gloria, Carla, Isabel, Sylvia, Evelyn, Doris, Arthur, Raymond, Harold, Lawrence, Neil, Brenda, Tracy, Simon, Wendy, Zoe, Ethan, Calvin, Sean, Ruth, Sheila, Miriam, Lorraine, Fay, Sophie

\paragraph{clothes\_socks}
blue, red, yellow, green, purple, pink, orange, black, white, gray

\paragraph{clothes\_shirt}
blue, red, yellow, green, purple, pink, orange, black, white, gray

\paragraph{clothes\_pant}
blue, red, yellow, green, purple, pink, orange, black, white, gray

\paragraph{clothes\_hat}
blue, red, yellow, green, purple, pink, orange, black, white, gray

\paragraph{clothes\_gloves}
blue, red, yellow, green, purple, pink, orange, black, white, gray

\paragraph{clothes\_underwear}
blue, red, yellow, green, purple, pink, orange, black, white, gray

\paragraph{hair}
blue, red, yellow, green, purple, pink, orange, black, white, gray

\paragraph{recent\_eat}
pizza pasta burrito sushi taco burger toast egg banana potatoes

\paragraph{recent\_listen}
rock, pop, country, electronic, folk, jazz, blues, classical, funk, ska, rap, synth, disco, reaggea

\paragraph{recent\_watch}
drama, comedy, thriller, romance, adventure, horror, sci-fi, action, western, fantasy, documentary, mystery, crime, musical

\paragraph{recent\_read}
fiction, mystery, novel, thriller, biography, sci-fi, non-fiction, essay, encyclopedia, dictionary

\paragraph{location} 
bathroom livingroom kitchen basement toilet balcony garden pool bedroom store university farm office bank tree museum school airport zoo train bus park butcher library restaurant mall mountain tunnel church river pond harbor taxi gallery bar pizzeria beach gym elevator insurance embassy police hospital festival monument laboratory observatory valley motorway viewpoint synagogue factory castle cave stadium arena cabin plaza amphitheater bridge pier vineyard forest cliff desert creek bay lighthouse orchard resort camp inn motel aquarium bazaar chapel monastery lookout campground retreat dock depot consulate manor theatre cathedral casino lodge mill bakery spa station diner gazebo terrace arcade boardwalk winery hill plateau ridge port oasis market fairground quarry mine grove auditorium cemetery dunes courthouse prison fort granary ranch promenade coliseum field tower pavilion silo bistro labyrinth cafe saloon brewery carnival marina estate safari cottage courtyard waterpark island greenhouse meadow lagoon ford hacienda village marketplace grotto maze golfcourse atrium academy waterfront peninsula cove summit plains